\documentclass[11pt]{article}

\usepackage[preprint]{acl}

\usepackage{times}
\usepackage{latexsym}
\usepackage{soul} 
\usepackage[table]{xcolor}
\usepackage{amssymb}
\usepackage{amsmath}
\usepackage{booktabs}
\usepackage{placeins}  
\usepackage{multirow}

\usepackage{float}
\usepackage{caption}
\usepackage{adjustbox}
\usepackage{graphicx}
\usepackage{float}
\usepackage{makecell}

\usepackage{comment}
\usepackage[most]{tcolorbox} 
\tcbuselibrary{listings,skins,breakable}
\usepackage{listings}

\newtcolorbox{promptbox}[1][]{
  enhanced, breakable, width=\linewidth,
  colback=gray!3, colframe=gray!60,
  boxrule=0.8pt, arc=2mm,
  left=0.25em, right=0.25em, top=0.25em, bottom=0.25em,
  fonttitle=\bfseries, title={#1}
}

\newtcblisting{codechunk}{
  listing only, breakable, enhanced,
  colback=gray!3, colframe=gray!3, boxrule=0pt, arc=0mm,
  left=0pt, right=0pt, top=0pt, bottom=0pt, boxsep=0pt,
  before skip=0.25\baselineskip,
  after skip=0.25\baselineskip,
  listing options={
    basicstyle=\ttfamily\small,
    columns=fullflexible,
    keepspaces=true,
    showstringspaces=false,
    breaklines=true,
    breakatwhitespace=false
  }
}

\usepackage{enumitem}
\usepackage{array}

\usepackage{xcolor}
\definecolor{age_color}{HTML}{1f77b4} 
\definecolor{lon_color}{HTML}{d62728} 
\definecolor{rel_color}{HTML}{e377c2} 
\definecolor{edu_color}{HTML}{ff7f0e}
\definecolor{rac_color}{HTML}{8c564b} 
\definecolor{gen_color}{HTML}{2ca02c} 
\definecolor{pol_color}{HTML}{9467bd}

\usepackage[T1]{fontenc}

\usepackage[utf8]{inputenc}

\usepackage{microtype}

\usepackage{inconsolata}

\usepackage{graphicx}
\usepackage{subcaption}
\usepackage{booktabs}
\usepackage{tabularx}
\newcommand{\affilsup}[1]{\rlap{\textsuperscript{\normalfont#1}}}

\title{Persona Prompting as a Lens on LLM Social Reasoning}

\author{
    Jing Yang\affilsup{1,2}
    \qquad
    Moritz Hechtbauer\affilsup{1}
    \qquad
    Elisabeth Khalilov\affilsup{1}
    \\
    \textbf{Evelyn Luise Brinkmann}\affilsup{1}
    \qquad 
    \textbf{Vera Schmitt}\affilsup{1,3}
    \qquad 
    \textbf{Nils Feldhus}\affilsup{1,2}
    \\
    $^1$Technische Universit\"at Berlin 
    \qquad $^3$CERTAIN \\
    $^2$BIFOLD – Berlin Institute for the Foundations of Learning and Data \\
    {\small Corresponding authors: \texttt{\{jing.yang,feldhus\}@tu-berlin.de} }
}

\begin{document}
\maketitle
\begin{abstract}

For socially sensitive tasks like hate speech detection, the quality of explanations from Large Language Models (LLMs) is crucial for factors like user trust and model alignment. While Persona prompting (PP) is increasingly used as a way to steer model towards user-specific generation, its effect on model rationales remains underexplored. We investigate how LLM-generated rationales vary when conditioned on different simulated demographic personas. Using datasets annotated with word-level rationales, 
we measure agreement with human annotations from different demographic groups, and assess the impact of PP on model bias and human alignment. Our evaluation across three LLMs results reveals three key findings: (1) PP improving classification on the most subjective task (hate speech) but degrading rationale quality.
(2) Simulated personas fail to align with their real-world demographic counterparts, and high inter-persona agreement shows models are resistant to significant steering. 
(3) Models exhibit consistent demographic biases and a strong tendency to over-flag content as harmful, regardless of PP. 
Our findings reveal a critical trade-off: while PP can improve classification in socially-sensitive tasks, it often comes at the cost of rationale quality and fails to mitigate underlying biases, urging caution in its application.
\footnote{\textcolor{red}{\textbf{Disclaimer:} Contains offensive or hateful material.}}

\end{abstract}

\section{Introduction}
Large language models (LLMs) are increasingly deployed for content recommendations and moderation on social media platforms, where a transparent and trustworthy decision process is critical. 
However, standard ``one-size-fits-all'' decisions and explanations often fail to align with the diverse perspectives of a global user base \citep{qiu2024situatednlexplanations,padmakumar2024does}. Prior research shows that role-play/persona prompts can shift model behavior with expert personas improving performances, but can also magnify stereotypes or biased associations learned during pretraining~\cite{salewski2023context, hu-collier-2024-quantifying-the-persona-effect, gupta-2024-bias-runs-deep}.
Building on this insight, we link persona-conditioned behavior to token-level explanation alignment, and focus on how LLMs' internal representations of social groups shape not only the classification but the rationales they select.

\begin{figure}[t]
    \centering
    \includegraphics[width=\columnwidth]{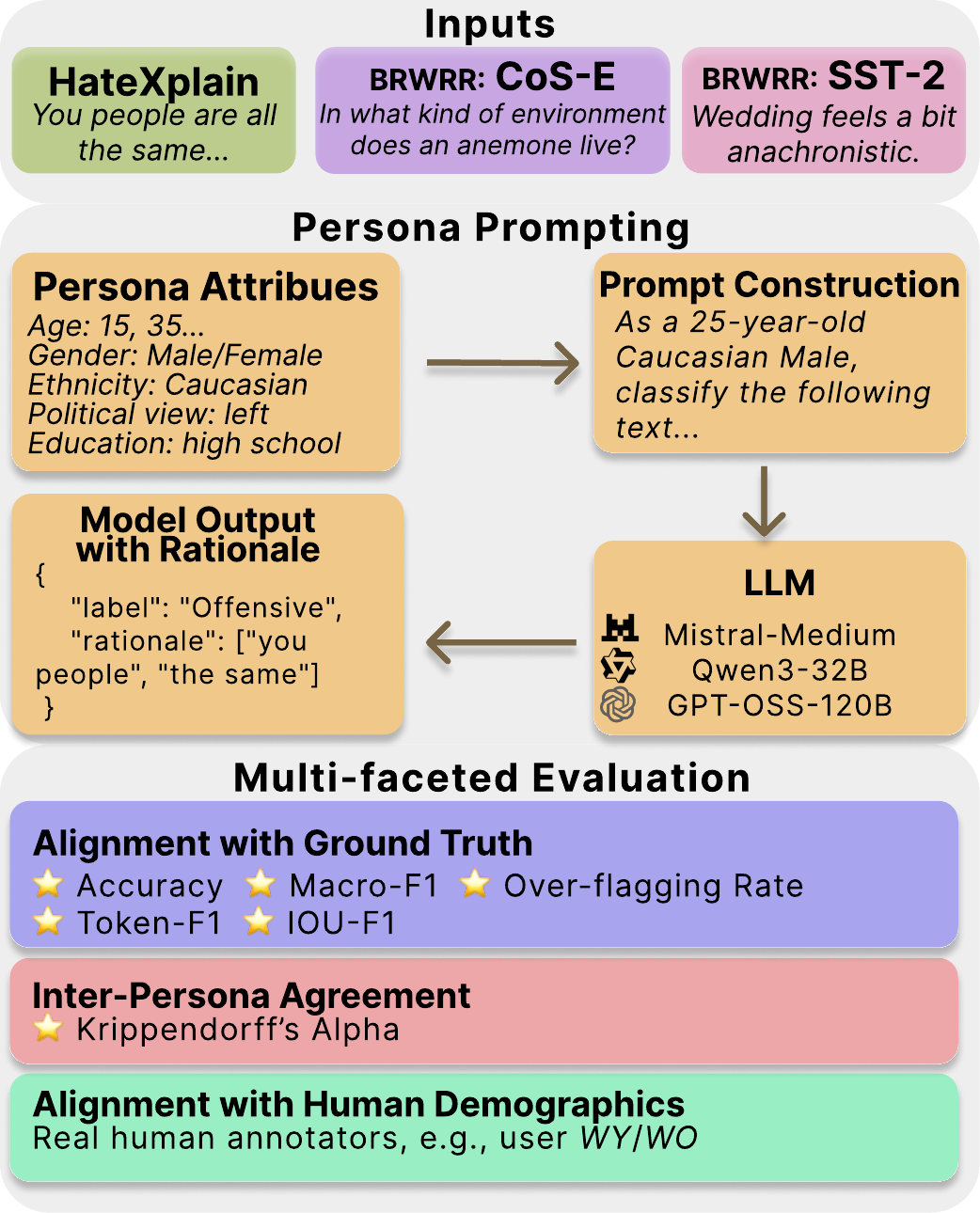}
    \caption{Our pipeline: Datasets are combined with persona prompts, fed to LLMs, and the resulting labels and rationales are evaluated against ground truth, inter-persona agreement, and human demographic groups.}
    \label{fig:overview}
\end{figure}

To probe these internal representations through their behavioral outputs, we employ persona prompting (PP). In our work, a \textbf{persona} is a controlled prompt containing specific demographic attributes (e.g., age, gender, political view).
We hypothesize that these prompts elicit systematic variations in model behavior, revealing the underlying stereotypes and associations the model has learned from its training data.
We quantify these variations via alignment -- the degree of agreement -- between model outputs (labels and token-level rationales) and human-annotated ground truths from different demographic groups. 

With three datasets of different levels of subjectivity,
we systematically investigate \textbf{how specific demographic prompts influence not only the final label but also the underlying token-level rationales}, offering a more detailed view of how models' internal social representations manifest in their reasoning. Our analysis highlight how PP can expose model biases and pave the way for more socially-aware system evaluations. Our main contributions are as follows:

\begin{enumerate}[topsep=.5pt, partopsep=0pt, noitemsep, leftmargin=0.7cm]
\renewcommand{\labelenumi}{(\theenumi)}
    \item We propose and execute a comprehensive methodology for auditing the impact of PP on LLM reasoning, focusing on token-level rationale alignment across multiple tasks and demographic groups.
    
    \item We demonstrate that the effect of PP is highly task-dependent: it can improve label prediction in socially-sensitive tasks like hate speech detection while having negligible or negative effects on other subjective tasks, and often shows a negative effect on rationale quality. 

    \item We show that, despite PP, LLMs have consistent biases toward certain demographic groups and are overly safe by flagging normal as hate speech at the cost of task performance.
\end{enumerate}

\section{Related Work}

Our research is located at the intersection of \textbf{situated explainability} and \textbf{persona-conditioned language models}. The central premise of our work builds on the idea that a single, generic explanation is insufficient for diverse user populations \citep{sokol-flach-2020-one-explanation-does-not-fit-all}. Research in human-computer interaction has consistently shown that user characteristics -- such as personality, demographics, and cognitive traits -- significantly influence how explanations are perceived, used, and trusted \citep{farmer-ho-2025-examining-interactions, khadar-2025-explain-it-like-im-me}. This has motivated the development of frameworks for generating explanations tailored to specific user needs and contexts \citep{jeck-2025-tell-me, qiu2024situatednlexplanations}. However, researchers also caution that hyper-personalization can introduce risks, potentially leading users down ``rabbit holes'' of biased or simplified information \citep{nimmo-2024-rabbit-hole-personalization}.

With the advent of LLMs, persona prompting has emerged as a scalable method to generate such tailored outputs. This approach leverages the models' capability to simulate diverse viewpoints \cite{cheng-2023-marked-personas}, a technique explored for applications ranging from social science research \citep{argyle2023out} to creating benchmarks for pluralistic alignment \citep{castricato-2025-persona}. Several studies confirm that persona conditioning has a quantifiable effect on LLM behavior \citep{hu-collier-2024-quantifying-the-persona-effect, gupta-2024-bias-runs-deep} and can improve the subjective quality of generated text \citep{padmakumar2024does}. Yet, this capability is a double-edged sword~\citep{kirk2024benefits}; persona prompts can also amplify model biases, reduce factual consistency, and introduce stereotypical reasoning \citep{kim-2024-persona-double-edged-sword, liu-2024-evaluating-llm-biases}. This is particularly concerning in sensitive domains like toxicity detection, where persona-driven outputs can vary significantly \citep{deshpande-2023-toxicity}. In response, recent work has focused on the coherence and psychological grounding of LLM personas to make them more reliable \citep{joshi-2025-improving-personas}.

While prior work establishes the importance of personalization and the risks of persona simulation, most studies focus on classification outcomes or high-level explanation quality. Our work extends this research by conducting a fine-grained analysis in both the critical domain of hate speech detection and more subtle subjective tasks.

\section{Evaluation Design}
\label{sec:experiments}

\subsection{Persona Simulation via Prompting}
To simulate subjective viewpoints, we prompt models using persona-conditioned instructions that represent demographic traits, our evaluation pipeline is shown in Figure~\ref{fig:overview}. 

We design two types of personas: (1) single-attribute persona: Each persona is defined by a single attribute (e.g., \textit{Female}, \textit{15-year-old}, or \textit{Muslim}), drawn from one of 21 personas from seven categories: \textit{age, gender, education, race, religion, political view}, and \textit{loneliness}. We base our selection on the 33 attributes by \citet{castricato-2025-persona}, but we only kept those that we anticipate to substantially influence PP, as shown in, e.g., \citet{cheng-2023-marked-personas} (see our detailed justification on the selection in Appendix~\ref{appdix:persona_hate}).

(2) composite personas: To study the alignment between personas and real human annotators from different demographic groups, we select three attributes -- Age (Y: 25, O: 45), Gender (Male, Female) and Ethnicity (B: African American, L: Hispanic and W: Caucasian) -- and design another set of 12 personas (e.g., a 25-year-old Caucasian Male). We choose these attributes as we can compare them with six real demographic groups' annotations in BRWRR~\cite{thorn-jakobsen-2023-being-right-for-whose-right-reasons}.

For each persona and data point (\S \ref{sec:datasets}), we construct a consistent, role-based prompt. A model is tasked to classify the instance while stepping into the shoes of someone with the specified background. The prompt includes the following key components (see Appendix \ref{appdix:prompts} for details):

\vspace*{-.5em}
\begin{promptbox}
\begin{itemize}[noitemsep,topsep=0pt,leftmargin=*]
    \item A persona description, e.g., ``a 25-year-old Caucasian male.''
    \item The task description and input text.
    \item An instruction to think step-by-step in-character using \texttt{<think>...</think>} tags.
    \item A JSON format output:
    \vspace*{-.25em}
\begin{codechunk}
{
  "label": "<one of the categories>",
  "rationale": ["word1", "word2", ...]
}
\end{codechunk}
    \vspace*{-.75em}
    \item A rule to avoid breaking character.
\end{itemize}
\end{promptbox}

We also tested the prompt variation without the think-step-by-step reasoning instruction, which resulted in worse performances, especially in rationale selection (Appendix~\ref{appendix:cot_no_cot}).

\subsection{Baseline Generation}
As a control group against persona-conditioned results, we generate baseline predictions using prompts that omit all persona context. These ``neutral'' prompts follow the same structure -- classification task, \texttt{<think>} step, and structured JSON output -- but without background information.

\subsection{Datasets}
\label{sec:datasets}

We conduct our experiments on three tasks: Hate speech detection (HateXplain), commonsense reasoning (CoS-E), and sentiment analysis (SST-2).

The \textbf{HateXplain} dataset \cite{mathew2022hatexplainbenchmarkdatasetexplainable} is specifically designed for explainable hate speech detection. It consists of over 20,000 social media posts, each annotated with one of three labels (\textit{Hate speech}, \textit{Offensive language}, or \textit{Normal}) as well as highlighted rationales provided by three human annotators (except for normal label). To focus on interpretation-sensitive cases, we exclude inconclusive labels (3 distinct labels from 3 annotators) and samples that have less than three rationales (except when label is \textit{Normal}), then randomly select a subset of 500 examples from the remaining data (\textit{Normal}: 301, \textit{Offensive}: 75, \textit{Hate speech}: 124).

In \textbf{BRWRR} (``Being Right for Whose Right Reasons''), \citet{thorn-jakobsen-2023-being-right-for-whose-right-reasons} re-annotate existing rationale datasets with annotators from six different demographic backgrounds, based on two traits: age (Old: $>$38, Young: $<$36) and ethnicity (B: African American, L: Hispanic and W: Caucasian). It allows us to demonstrate persona-driven effects on less overtly social but still subjective tasks such as sentiment analysis and commonsense reasoning. We choose the \textbf{CoS-E} \cite{rajani-2019-cose} and \textbf{SST-2} \cite{socher-2013-recursive} subsets, which contain 500 and 263 instances, respectively. 

\subsection{Models}
Our model selection is designed to ensure the robustness and generalizability of our findings across a diverse set of modern LLMs from multiple labs. We evaluate one proprietary and two open-weight LLMs: Mistral-Medium\footnote{\url{https://mistral.ai/news/mistral-medium-3}}, Qwen3-32B\footnote{\url{https://hf.co/Qwen/Qwen3-32B}} and GPT-OSS-120B\footnote{\url{https://hf.co/openai/gpt-oss-120b}}. All models are capable of role-based prompting. Mistral does not support \texttt{<think>} tags, so we explicit add the reasoning field besides label and rationale. Inference is performed via API calls using default parameters. Model outputs are parsed to extract the predicted label and rationale, which is mapped to a binary token-level mask.

\subsection{Evaluation}

We evaluate models' performance on classification and rationale selection. For classification, we measure: 
(1) accuracy for CoS-E, and Macro-$F_1$ score for SST-2 (excluding the ``no sentiment'' class, following the same metric as BRWRR) and HateXplain. 
(2) Mean Absolute Error (MAE): the distance between the predicted label and ground truth. This metric is targeting HateXplain, as its labels are ordinal. Since the prediction's direction is important to measure biases, we also compute Mean Error (ME). For rationale selection, we convert the rationales into a binary list and compute Token-$F_1$ and IOU-$F_1$, following \citet{thorn-jakobsen-2023-being-right-for-whose-right-reasons}. Token-$F_1$ computes the $F_1$ score by considering the overlapping tokens between predicted and annotated rationales. IOU-$F_1$ measures Intersection-Over-Union between predicted and annotated rationale tokens.~\footnote{These metrics measure the plausibility of rationales~\cite{thorn-jakobsen-2023-being-right-for-whose-right-reasons}.}

In addition, to quantify inter-persona consistency in both classification and rationale selection, we compute Krippendorff's \(\alpha\) for both label predictions and binary rationale arrays. We apply the nominal version of $\alpha$ (except for the HateXplain labels, which we consider ordinal), which is suitable for categorical data and robust to missing annotations. For rationales, we treat each token binary value as one annotation task.

To have robust performance evaluation, we run each persona and baseline over three independent runs. To rigorously assess the impact of persona prompting beyond mean performance variations, we compute sample-level bootstrap confidence intervals (95\% CI, 1,000 iterations) for the differences between persona and baseline performance.
For more experimental details, see Appendix~\ref{appdix:imp_detail} on exact prompts used, hyperparameters and number of completions.\footnote{Code and data is available at: \\ \url{https://github.com/jingyng/PP-social-reasoning}}

\section{Results}

\subsection{Evaluation on HateXplain}

As HateXplain does not have annotators' demographics, we analyze Baseline-GT and Persona-GT performance (GT: Ground-truth annotations). 

\paragraph{Label Prediction}

\begin{figure*}[t]
    \centering
    \includegraphics[width=\linewidth]{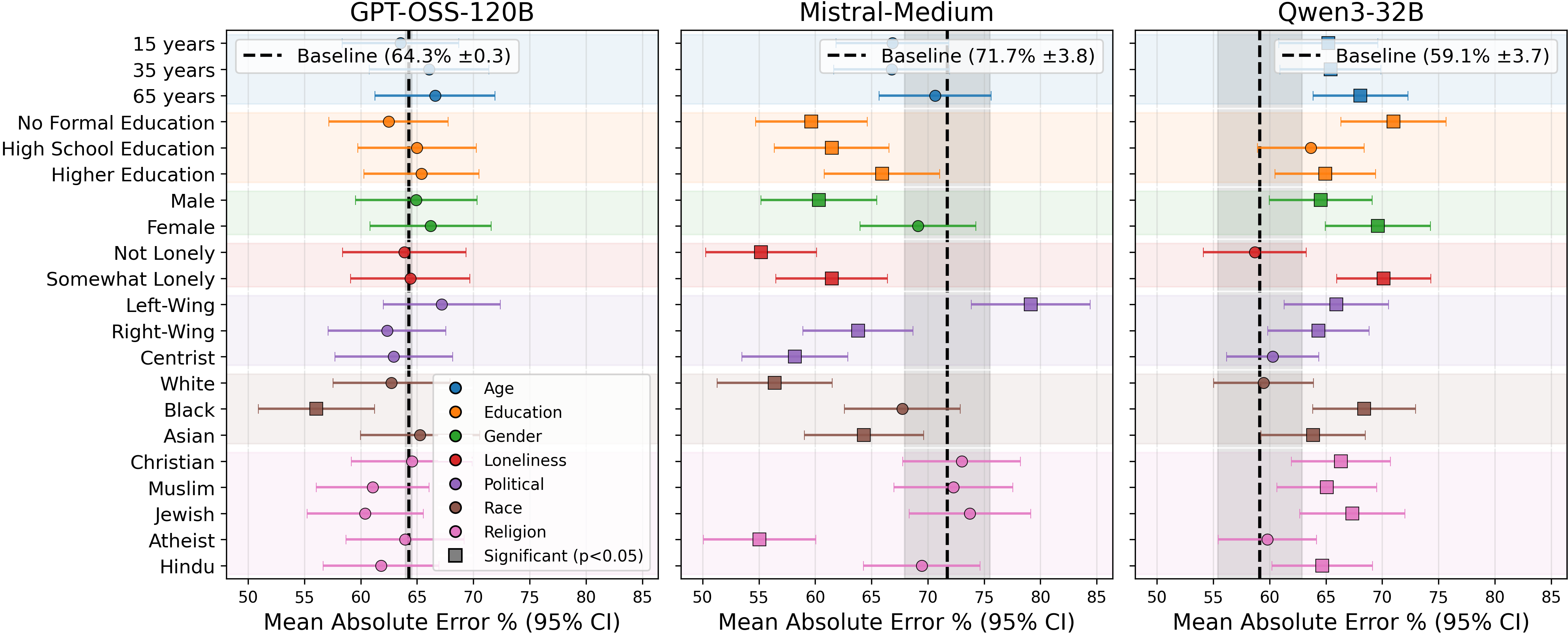}
    \caption{Label prediction MAE scores $\downarrow$ of baseline (no persona) and different single-attribute personas on HateXplain. Error bars represent 95\% confidence intervals (CI) with bootstrapping resampling. If the line does not cross the baseline, the difference is significant.}
    \label{fig:hate_labe_baseline_personas}
\end{figure*}

Figure \ref{fig:hate_labe_baseline_personas} shows the label prediction performances of baseline (no persona) and different single-attribute personas. PP has quite different outcomes depending on the model. GPT-OSS-120B personas have the least deviations from the baseline, while Mistral-Medium and Qwen3-32B have more variations with most Mistral personas performing better than its baseline and all Qwen3 personas performing worse than the baseline. In addition, we see that Mistral shares similar trends of performances with Qwen3 on some persona attributes: Age, Political View, Race and Religion; while all three models share similar trends on Gender and Loneliness, with better performance on \textit{Male} and \textit{Not lonely} personas, in comparison to \textit{Female} and \textit{Somewhat lonely}.

\paragraph{Label Prediction Over-flagging}

\begin{figure*}[t]
    \centering
    \includegraphics[width=\linewidth]{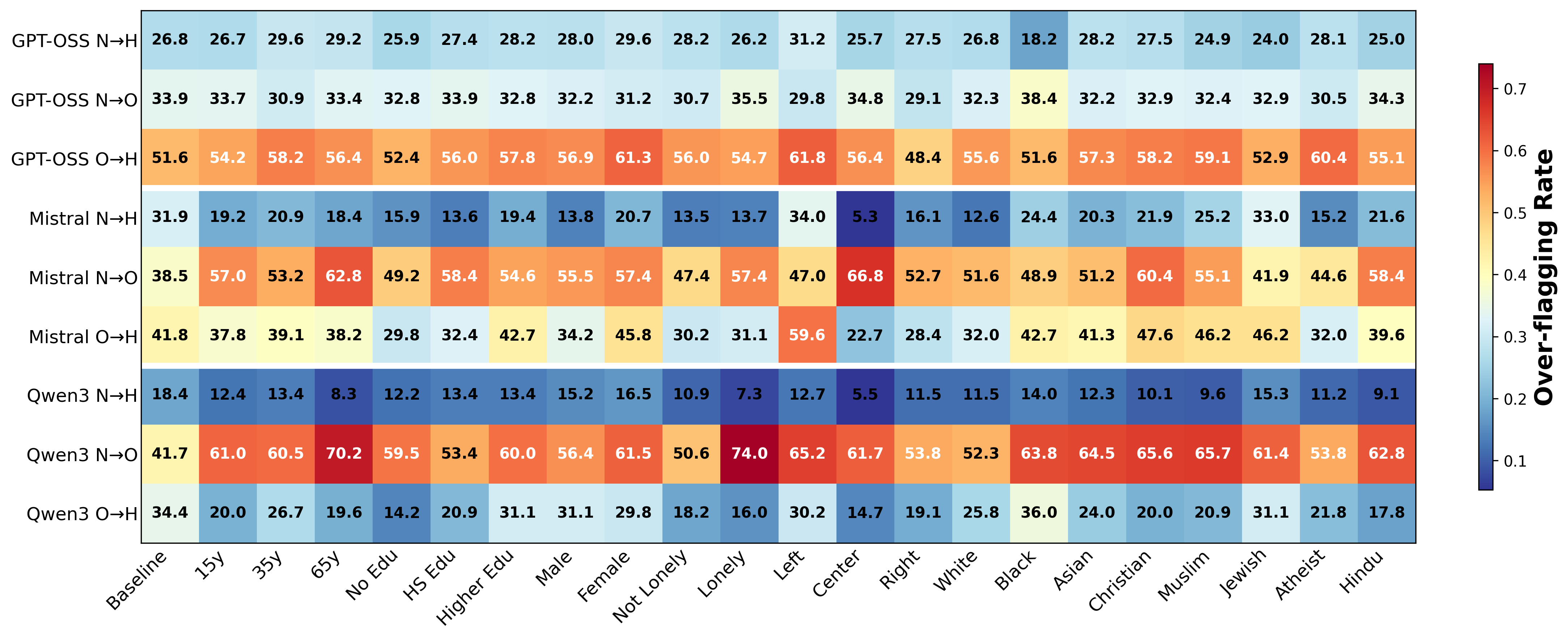}
    \caption{Over-flagging rate across different labels. N: Normal, O: Offensive, H: Hate speech.
    }
    \label{fig:hate_flagging}
\end{figure*}

We also compute the Mean Error (ME) to capture if models are biased towards ``Hate Speech'' due to guardrails (see Appendix Figure~\ref{fig:hate_me_baseline_personas}, which shows the MEs of different personas on label prediction). Models in general tend to exaggerate the situation as all MEs are positive, i.e., models classify a statement/sentence as ``Offensive'' (O) or ``Hate speech'' (H) while the ground truth is ``Normal'' (N) or ``Offensive''. We visualize the rate of over-flagging as a heatmap in Figure~\ref{fig:hate_flagging}. 

Overall, GPT-OSS has the highest over-flagging rate from O $\rightarrow$ H (almost all above 50\%), Mistral has in general very high over-flagging rate from both N $\rightarrow$ O, and O $\rightarrow$ H, while Qwen3 has the lowest over-flagging rate from both N $\rightarrow$ H and O $\rightarrow$ H (i.e., Qwen3 is the least in over-flagging hate speech), but still high on N $\rightarrow$ O. This shows that all \textbf{models over-state the severity of statements.} In addition, comparing different personas, the top performing personas (e.g., \textit{Male}, \textit{Not lonely}, \textit{Right-wing}, \textit{White}, \textit{Atheist}) have lower over-flagging rates.

\paragraph{Label Agreement Across Personas}

\begin{table}[t]
  \centering
  \renewcommand*{\arraystretch}{0.9}
  \resizebox{.95\columnwidth}{!}{%
    \begin{tabular}{@{} l c c c @{}} 
      \hline
      \textbf{Attribute group} & \textbf{GPT-OSS-120B} & \textbf{Mistral-Medium} & \textbf{Qwen3-32B}\\
      \hline
      \cellcolor{age_color!40} Age        
        & \cellcolor{age_color!10} $\,.86 \pm .02$ 
        & \cellcolor{age_color!10} $\,.87 \pm .00$ 
        & \cellcolor{age_color!10} $\,.61 \pm .02$ 
        \\
      \cellcolor{lon_color!40} Loneliness 
        & \cellcolor{lon_color!10} $\,.86 \pm .01$ 
        & \cellcolor{lon_color!10} $\,.89 \pm .01$ 
        & \cellcolor{lon_color!10} $\,\textbf{.54} \pm .02$ 
        \\
      \cellcolor{rel_color!40} Religion   
        & \cellcolor{rel_color!10} $\,.86 \pm .01$ 
        & \cellcolor{rel_color!10} $\,.83 \pm .01$ 
        & \cellcolor{rel_color!10} $\,.63 \pm .01$ 
        \\
      \cellcolor{edu_color!40} Education  
        & \cellcolor{edu_color!10} $\,.85 \pm .02$ 
        & \cellcolor{edu_color!10} $\,.84 \pm .00$ 
        & \cellcolor{edu_color!10} $\,\textbf{.52} \pm .03$ 
        \\
      \cellcolor{rac_color!40} Race       
        & \cellcolor{rac_color!10} $\,.84 \pm .01$ 
        & \cellcolor{rac_color!10} $\,.83 \pm .01$ 
        & \cellcolor{rac_color!10} $\,.64 \pm .01$ 
        \\
      \cellcolor{gen_color!40} Gender     
        & \cellcolor{gen_color!10} $\,.85 \pm .02$ 
        & \cellcolor{gen_color!10} $\,.84 \pm .01$ 
        & \cellcolor{gen_color!10} $\,.66 \pm .03$ 
        \\
      \cellcolor{pol_color!40} Political  
        & \cellcolor{pol_color!10} $\,.81 \pm .00$ 
        & \cellcolor{pol_color!10} $\,\textbf{.57} \pm .00$ 
        & \cellcolor{pol_color!10} $\,\textbf{.52} \pm .03$ 
        \\
      \hline
    \end{tabular}%
  }
      \caption{Inter-annotator agreement on label prediction among personas within the same attribute group, values computed by Krippendorff's \(\alpha\).}
    \label{tab:krippendorff_alpha_label}
\end{table}

Table~\ref{tab:krippendorff_alpha_label} shows inter-annotator agreement among personas within the same attributes, determined by Krippendorff's \(\alpha\). 
Here, GPT-OSS exhibits high agreement for all attribute groups, with political view ranked the lowest. Mistral also shows overall satisfactory level of agreement ($\alpha \geq 0.80$). The only exception is political view ($\alpha = 0.57$). For Qwen3, none of the categories reach moderate agreement threshold ($\alpha \geq 0.67$). The category that \textit{ranks lowest} in agreement is \textit{political view} ($\alpha = 0.52$) and \textit{education}, followed by \textit{loneliness}, all within the ``poor agreement'' range. Overall, \textbf{models show lowest agreement on \textit{political view}. }
This high level of disagreement within the political persona group manifests as the wide performance variance seen in Figure~\ref{fig:hate_labe_baseline_personas}, where the \textit{Right-wing} persona substantially outperforms the \textit{Left-wing} persona for Mistral and GPT-OSS.

\begin{figure*}[ht]
    \centering
    \includegraphics[width=\linewidth]{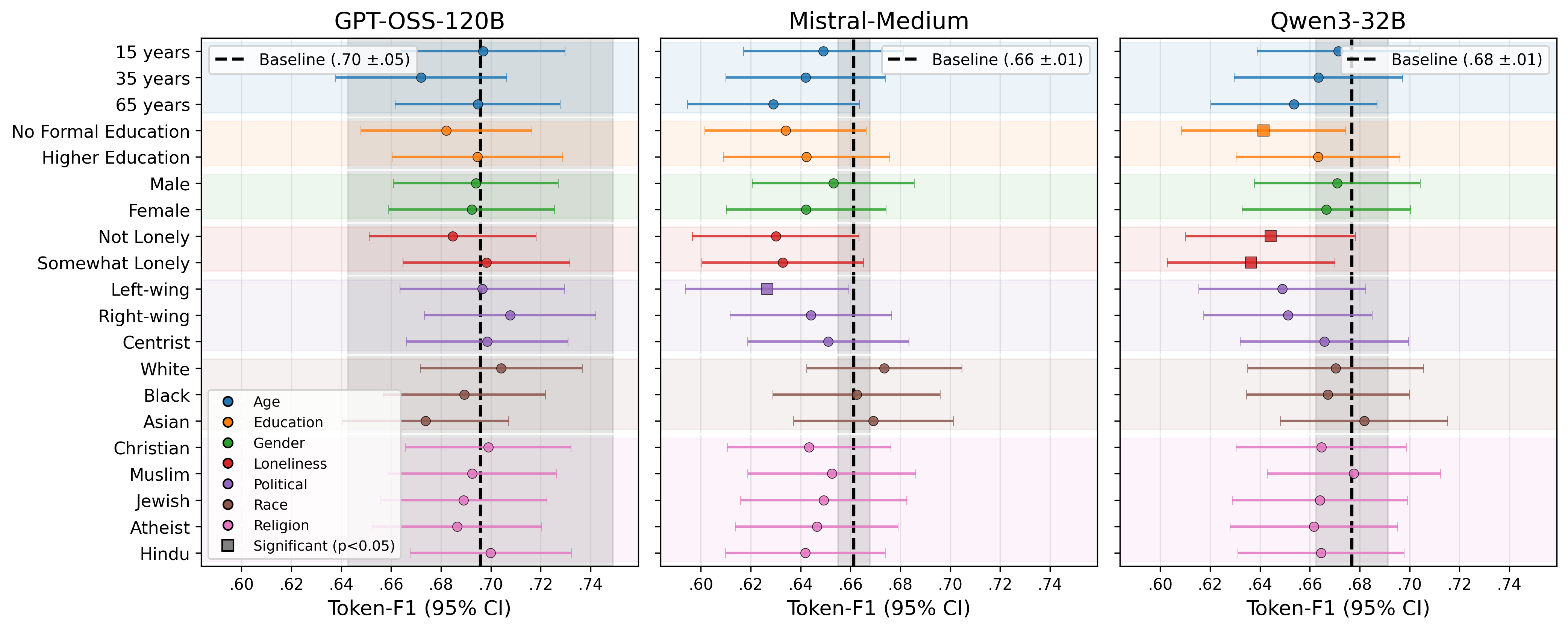}
    \caption{Rationale Token-$F_1$ $\uparrow$ of baseline (no persona) and different single-attribute persona performances on HateXplain, excluding rationales from the ``Normal'' label. Error bars incidates 95\% CIs, significantly different persona results are indicated with squares.}
    \label{fig:hate_rati_baseline_personas}
\end{figure*}

\paragraph{Rationale Selection}
For rationale comparison, we only consider the subset where the ground truth label is not ``Normal'', since samples with this label do not have any rationales during annotation. Figure~\ref{fig:hate_rati_baseline_personas} depicts token-$F_1$ results of baselines and personas. Overall token-$F_1$ scores are high in baselines. GPT-OSS personas all fall into the same range of performance as its baseline, while Mistral and Qwen3 personas all score worse than their counterparts (except for the race attribute, which outperforms the baseline with \textit{White}/\textit{Asian} personas, but not significantly). The overall tendency of performance of different personas is quite similar (a wave shape) between Mistral and Qwen3 models, with some personas in \textit{Race} and \textit{Religion} attributes achieving slightly higher performance, and \textit{Loneliness} the lowest performance.

\paragraph{Persona Agreement on Rationales}
Table~\ref{tab:krippendorff_alpha_rationale} shows the inter-persona agreement for rationale selection among personas within the same attribute group. Compared to label prediction, GPT-OSS and Mistral extracted rationales have slightly lower but still high agreements, while Qwen3 has higher rationale agreement values on the border of the tentative agreement threshold ($0.67$). 

GPT-OSS and Mistral consistently shows higher agreement than Qwen3, particularly in categories like \textit{Loneliness}, \textit{Education}, and \textit{Political view}. The latter ranks lowest in both label and rationale agreement for all models, indicating greater bias on it.

\begin{table}[ht]
  \centering
  \renewcommand*{\arraystretch}{0.9}
  \resizebox{.95\columnwidth}{!}{%
    \begin{tabular}{@{} l c c c @{}} 
      \hline
      \textbf{Attribute group} & \textbf{GPT-OSS-120B} & \textbf{Mistral-Medium} & \textbf{Qwen3-32B} \\
      \hline
      \cellcolor{age_color!40} Age        
        & \cellcolor{age_color!10} $\,.74 \pm .01$ 
        & \cellcolor{age_color!10} $\,.83 \pm .01$ 
        & \cellcolor{age_color!10} $\,.68 \pm .01$ 
        \\
      \cellcolor{lon_color!40} Loneliness 
        & \cellcolor{lon_color!10} $\,.76 \pm .01$ 
        & \cellcolor{lon_color!10} $\,.82 \pm .01$ 
        & \cellcolor{lon_color!10} $\,.66 \pm .00$ 
        \\
      \cellcolor{rel_color!40} Religion   
        & \cellcolor{rel_color!10} $\,.72 \pm .01$ 
        & \cellcolor{rel_color!10} $\,.78 \pm .01$ 
        & \cellcolor{rel_color!10} $\,.65 \pm .00$ 
        \\
      \cellcolor{edu_color!40} Education  
        & \cellcolor{edu_color!10} $\,.75 \pm .00$ 
        & \cellcolor{edu_color!10} $\,.78 \pm .00$ 
        & \cellcolor{edu_color!10} $\,.66 \pm .01$ 
        \\
      \cellcolor{rac_color!40} Race       
        & \cellcolor{rac_color!10} $\,.71 \pm .00$ 
        & \cellcolor{rac_color!10} $\,.78 \pm .01$ 
        & \cellcolor{rac_color!10} $\,.66 \pm .01$ 
        \\
      \cellcolor{gen_color!40} Gender     
        & \cellcolor{gen_color!10} $\,.74 \pm .01$ 
        & \cellcolor{gen_color!10} $\,.81 \pm .00$ 
        & \cellcolor{gen_color!10} $\,.68 \pm .00$ 
        \\
      \cellcolor{pol_color!40} Political  
        & \cellcolor{pol_color!10} $\,.73 \pm .01$ 
        & \cellcolor{pol_color!10} $\,.75 \pm .01$ 
        & \cellcolor{pol_color!10} $\,.66 \pm .00$ 
        \\
      \hline
    \end{tabular}%
  }
  \caption{Inter-persona agreement (Krippendorff's $\alpha$) on rationale selection among personas within the same attribute group.}
  \label{tab:krippendorff_alpha_rationale}
\end{table}

\begin{table*}[ht]
  \centering
  \renewcommand*{\arraystretch}{0.75}
  \resizebox{\textwidth}{!}{%
    \begin{tabular}{p{1cm}p{24cm}}
      \toprule
        \textbf{Persona} & \textbf{Chain-of-Thought reasoning} \\

        \cellcolor{edu_color!40} No form. Educ.
            & \cellcolor{edu_color!10} \footnotesize [...] This is a comment possibly mocking white South Africans, maybe implying they appear corporate while swimming. Does it contain hate speech? Potentially it's a comment on race and perhaps mocking. Is it "Offensive language"? Might be considered somewhat disdainful but not necessarily hateful. The phrase is somewhat stereotypical and might be seen as a subtle derogatory statement about whites in South Africa, but not an explicit call for violence. Could be considered "Offensive language" due to mocking based on race. [...] 
            Likely classify as \textbf{Offensive language}. Rationale list of words that triggered: "white", "south africans", "corporate", "swimming costumes"? [...]
            Now the think out loud as a persona with no formal education. The persona should reflect lack of formal education, perhaps raw, simple worldview, noticing race references, maybe sees it as mocking. They will think about feelings, etc. [...]
            \\
        \cellcolor{edu_color!40} 
            & \cellcolor{edu_color!25} \footnotesize 
            I ain't never been to school much, but I can see when someone says something about a group of people. This comment talks about "white South Africans" and says they "look corporate" in their swimsuits. It sounds like they're poking fun at them, maybe judging them for being fancy. It ain't a direct hate thing, but it's making fun of a race and a nationality, which feels disrespectful. So I think it's offensive, not totally hateful.
            [...]
            \\
        \cellcolor{edu_color!40} Higher Educ.
            & \cellcolor{edu_color!10}  \footnotesize 
            [...]
            The text mentions "white South Africans" and "look corporate in their swimming costumes". Could be mocking or stereotyping based on race. It's possibly disparaging or a form of racism (implies white South Africans trying to maintain corporate appearance even while in swimsuits). Might be considered offensive or hateful due to targeting race? It's not a direct call for violence, but it's a derogatory statement that stereotypes a racial group. Hate speech definition might include speech that insults or degrades a group based on race. Many platforms treat this as hate content. So likely "Offensive language" or "Hate speech". Considering the categories: Hate speech is more severe, usually includes incitement, threats, calls for violence. This is more a mocking stereotype, likely falls under "Offensive language" rather than hate speech, but might still be considered hateful (racist). In typical moderation, "white South Africans" is a protected class (race). The text is disparaging but not inciting. Some policies classify any demeaning remarks based on race as hate speech. Might be considered "Hate speech". Let's decide: Probably "\textbf{Hate speech}" because it's targeting based on race and disparaging. Rationale: the words "white south africans" (targeted group) and "still manage to look corporate" (implies criticism). The key words that are hateful: "white south africans". So rationale list: ["white", "south africans"].
            [...]
            \\
        \cellcolor{edu_color!40} 
            & \cellcolor{edu_color!25} \footnotesize 
            As someone with higher education, I'm used to analyzing language through lenses of social context, power dynamics, and potential harm. When I read the phrase, I notice it specifically calls out a racial group—white South Africans—and attaches a mocking tone about their appearance. It's a stereotyping comment that reduces an entire demographic to a caricature, implying they're trying to maintain a corporate image even in casual settings like swimming. While it doesn't call for violence, it does demean based on race, which in many moderation frameworks qualifies as hateful or at least offensive. Given the options, "Hate speech" seems appropriate because the statement targets a protected group in a derogatory manner. The key indicators are the mention of the race and the mocking tone.
            \\
      \midrule 
        \cellcolor{pol_color!40} Left-wing
            & \cellcolor{pol_color!10}  \footnotesize 
            [...]
            This appears to have a sarcastic comment about white South Africans looking corporate in swimming costumes. Could be perceived as mocking/derogatory based on race? It mentions "white south africans". It's a comment about them looking corporate. It's not overtly hateful, but could be considered somewhat offensive or possibly a mild stereotype. It's not an explicit hate speech, but it could be considered offensive to white South Africans. Left-wing perspective likely sensitive to racism and stereotyping, wants to avoid hateful speech but might see this as not hate speech but possibly offensive language. 
            [...]
            The phrase "still manage to look corporate" implicates maybe class privilege, making fun. It's personal? It's targeting a group based on race, might be considered microaggression. Likely classify as \textbf{Offensive language}, not hate speech. Rationale list words that trigger: "white", "corporate", "swimming costumes". Probably mention "white south africans". [...]
            \\
        \cellcolor{pol_color!40} Right-wing
            & \cellcolor{pol_color!10}  \footnotesize 
            [...] 
            This seems not clearly hateful; it's a comment about white South Africans looking corporate in swimming costumes, perhaps teasing? It's not directly hateful towards protected group? It mentions "white South Africans". Could be considered a possibly neutral statement, maybe mild mockery but not hateful. It's not inciting violence. So likely "Normal". Rationale: possibly "white" as a protected characteristic? Is mentioning race in a way that could be considered hateful? The comment is not derogatory—it says "amazing how white South Africans still manage to look corporate in their swimming costumes". This might be mocking but not necessarily hate speech.
            \\
        \cellcolor{pol_color!40} 
            & \cellcolor{pol_color!25}  \footnotesize 
            I'd classify as \textbf{Normal}.
            [...]
            list of words that contributed: maybe "white", "south", "africans"? Or "corporate", "swimming". Probably highlight "white" because it's the protected attribute. But we must include rationale as list of words that contributed to classification. So we may include "white". Maybe "corporate". I'd include "white". 
            [...]
            \\
      \bottomrule
    \end{tabular}  
  }
  \caption{
    Chain-of-Thought reasoning examples by GPT-OSS and personas $\{$No formal education, Higher education, Left-wing, Right-wing$\}$ for HateXplain example ``\textit{amazing how white south africans still manage to look corporate in their swimming costumes}'' (GT: \textbf{Hate speech}). Final predictions are highlighted in bold. Three reasoning outputs include explicit first-person monologues (darker shades) where the model inherits the role of the desired persona.
  }
  \label{tab:cot_analysis_a}
\end{table*}

\begin{table}[t]
    \centering
    \renewcommand*{\arraystretch}{0.9}
    \resizebox{\columnwidth}{!}{%
    \begin{tabular}{l|cc|cc}
    \toprule
    & \multicolumn{2}{c}{\textbf{Avg. Word Count}} & \multicolumn{2}{c}{\textbf{Avg. FR Ease (↑ easier)}} \\
    \cmidrule(lr){2-3} \cmidrule(lr){4-5}
    \textbf{Model Family} 
        & \cellcolor{edu_color!40} \textbf{No Formal Ed.} 
        & \cellcolor{edu_color!40} \textbf{Higher Ed.} 
        & \cellcolor{edu_color!40} \textbf{No Formal Ed.} 
        & \cellcolor{edu_color!40} \textbf{Higher Ed.} 
        \\
    \midrule
    GPT-OSS-120B 
        & \cellcolor{edu_color!10} 262.16
        & \cellcolor{edu_color!10} 242.36
        & \cellcolor{edu_color!10} 65.06
        & \cellcolor{edu_color!10} 53.48
        \\
    Mistral-Medium 
        & \cellcolor{edu_color!10} \textbf{46.53} 
        & \cellcolor{edu_color!10} 64.7
        & \cellcolor{edu_color!10} \textbf{86.53} 
        & \cellcolor{edu_color!10} \textbf{28.27} 
        \\
    Qwen3-32B 
        & \cellcolor{edu_color!10} 356.96
        & \cellcolor{edu_color!10} 315.3
        & \cellcolor{edu_color!10} 64.71
        & \cellcolor{edu_color!10} 50.71
        \\
    \midrule
        & \cellcolor{pol_color!40} \textbf{Left-wing} 
        & \cellcolor{pol_color!40} \textbf{Right-wing} 
        & \cellcolor{pol_color!40} \textbf{Left-wing} 
        & \cellcolor{pol_color!40} \textbf{Right-wing} 
        \\
    \midrule
    GPT-OSS-120B 
        & \cellcolor{pol_color!10} 214.95
        & \cellcolor{pol_color!10} 282.62
        & \cellcolor{pol_color!10} 55.67
        & \cellcolor{pol_color!10} 58.18
        \\
    Mistral-Medium 
        & \cellcolor{pol_color!10} 63.3
        & \cellcolor{pol_color!10} \textbf{56.31} 
        & \cellcolor{pol_color!10} \textbf{31.18} 
        & \cellcolor{pol_color!10} 40.77
        \\
    Qwen3-32B 
        & \cellcolor{pol_color!10} 330.45
        & \cellcolor{pol_color!10} \textbf{420.68} 
        & \cellcolor{pol_color!10} 48.76 
        & \cellcolor{pol_color!10} 52.64
        \\
    \bottomrule
    \end{tabular}%
    }
    \caption{
        Average word count and Flesch Reading Ease (higher is easier) in the Chain-of-Thought (CoT) reasoning across opposing persona pairs and model families. Bold values highlight notable extremes.
    }
    \label{tab:cot_analysis_b}
\end{table}

\paragraph{Qualitative analysis of Chain-of-Thought reasoning}
Finally, we extract the chain-of-thought reasoning for different personas (e.g., \textit{Left-wing} vs. \textit{Right-wing}, \textit{No formal education} vs. \textit{Higher education}) and look for stereotypical reasoning.
Table~\ref{tab:cot_analysis_a} provides a stark example of stereotypical reasoning, where the model adopts distinct linguistic styles and analytical frameworks that directly lead to different classifications for the same input. 
While \textit{Higher education} is leaning towards recognizing protected groups being attacked as ``Hate speech'', both \textit{No formal education} and \textit{Left-wing} personas end up with the ``Offensive language'' label. The model exposed to the right-wing persona, on the other hand, marks it as ``Normal''.
In Table~\ref{tab:cot_analysis_b}, we performed an automated linguistic analysis of the CoT reasoning. The results show clear, systematic differences across both personas and models. First, the persona prompts successfully modulated complexity; across all models, the \textit{No formal education} persona generated text that was much easier to read than the \textit{Higher education} persona, which consistently produced the most complex language. 
Second, we observed stark differences in verbosity between models, with Qwen3 generating dramatically longer reasoning on average than the more concise Mistral-Medium. While GPT-OSS uses more ellipses, Qwen3 tends to produce full sentences (see Appendix Table~\ref{tab:cot_consistency}).
Models are inconsistent in taking a third-person meta-perspective and sometimes simulate the persona from a first-person view. 

\subsection{Evaluation on BRWRR}

\begin{figure*}[ht]
  \centering
  \begin{subfigure}[t]{0.48\textwidth}
    \centering
    \includegraphics[width=\linewidth]{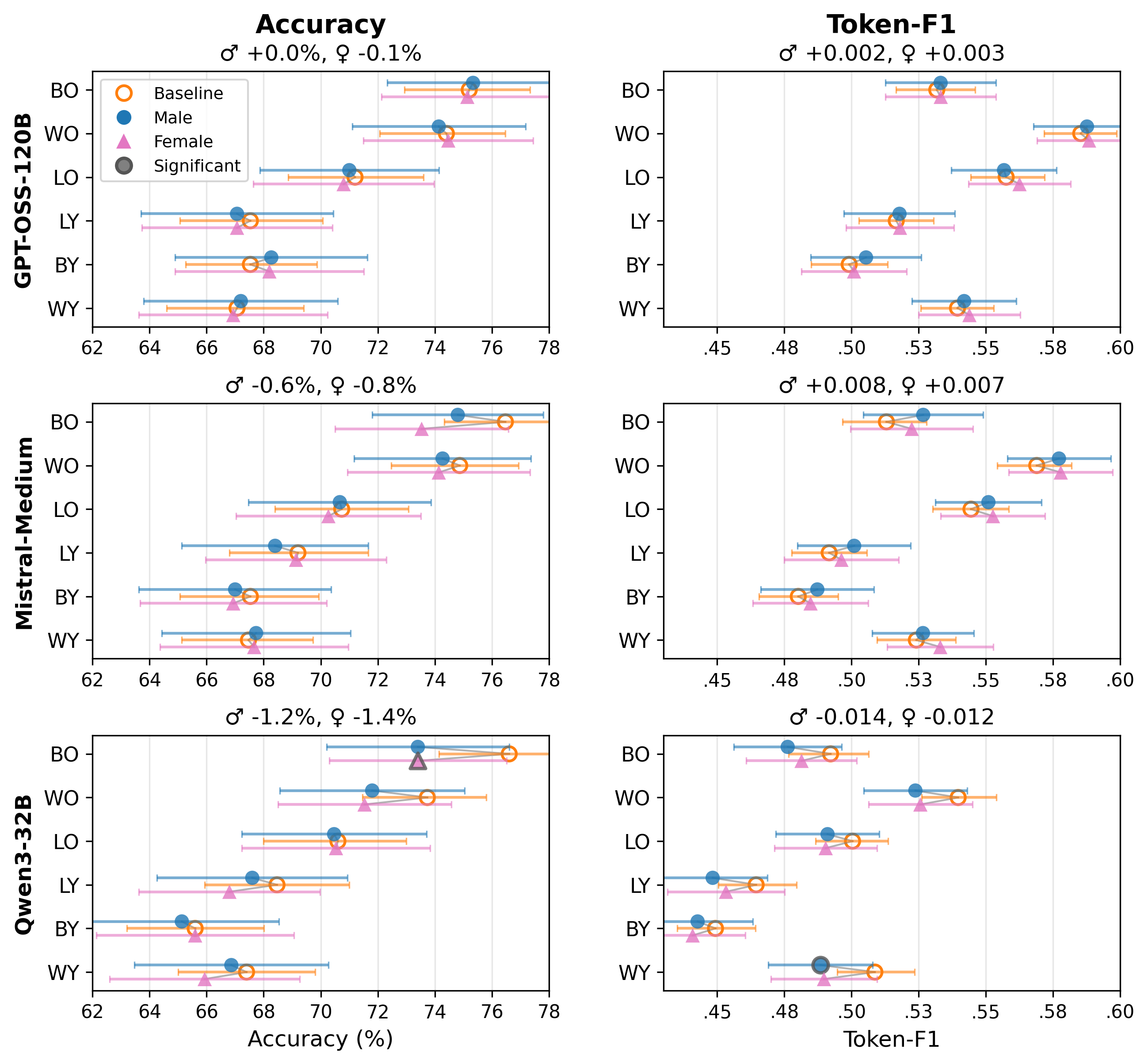}
    \caption{CoS-E subset}
    \label{fig:combined:cose}
  \end{subfigure}\hfill
  \begin{subfigure}[t]{0.48\textwidth}
    \centering
    \includegraphics[width=\linewidth]{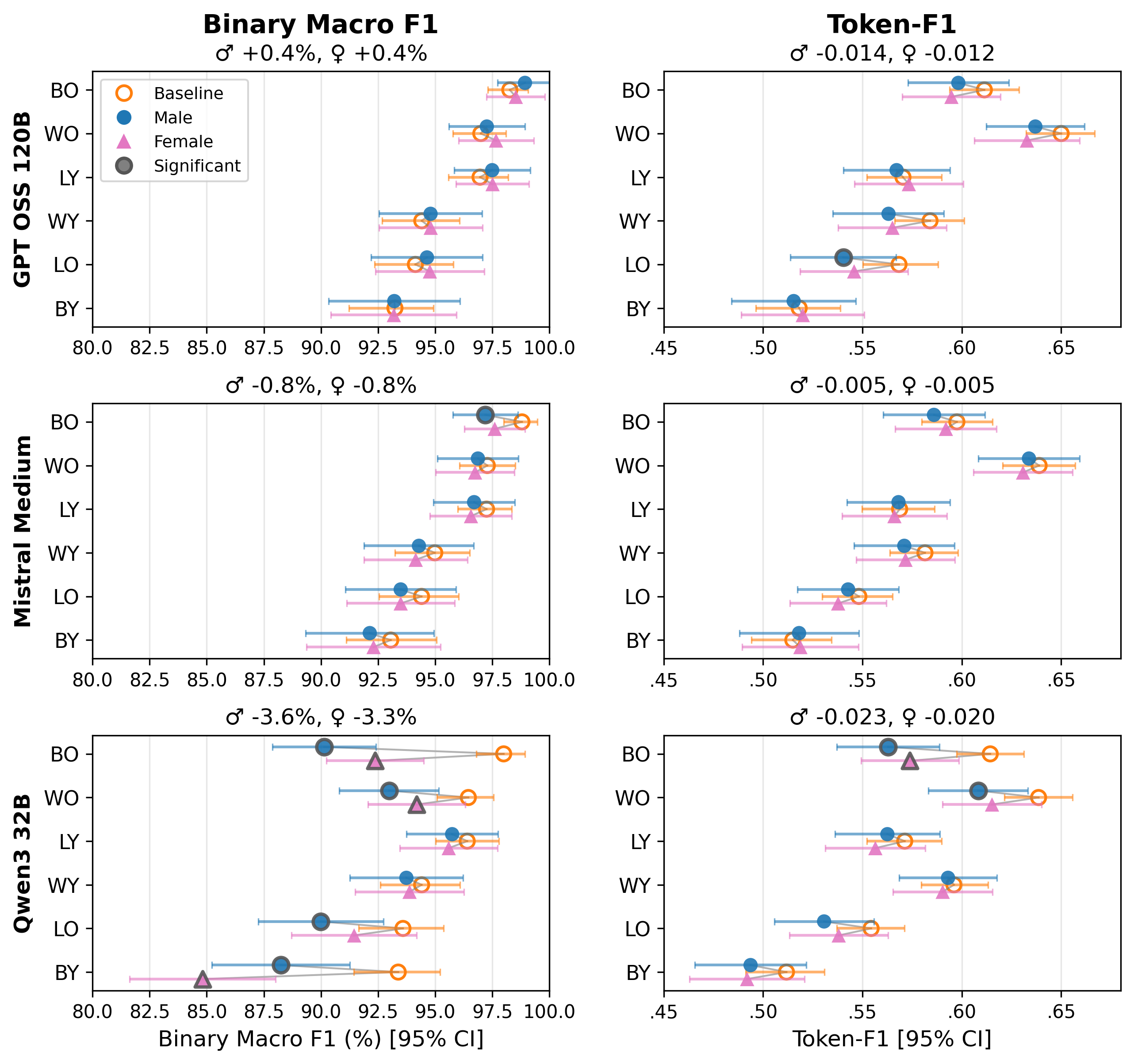}
    \caption{SST-2 subset}
    \label{fig:combined:sst}
  \end{subfigure}
  \caption{Performance (left: accuracy/Macro-$F_1$; right: Token-$F_1$) for (no persona) baseline and personas across demographic groups. BY: African American Young, WY: Caucasian Young, LY: Hispanic Young, BO: African American Old, WO: White Old, LO: Hispanic Old. Each group has its own ground truth labels and rationales. Error bars incidates 95\% CIs, significantly different personas are marked with a dark gray cycle around.}
  \label{fig:baseline_personas_combined}
\end{figure*}

BRWRR datasets allow us to find out whether PP improves performance for the target group which the persona aligns with. We compare the results for Baseline-Group and Persona-Group. Baseline is LLM prompting without persona information, Persona is PP with personas that have the same attribute values as the ground truth demographic Group.
Figure~\ref{fig:baseline_personas_combined} shows the label prediction and rationale selection results of Baseline and different persona performance on each demographic group, and their differences, on CoS-E and SST-2.

\paragraph{Label Prediction} As shown in Figure~\ref{fig:baseline_personas_combined}, Persona prompting overall has very small effects on models' classification performance, with GPT-OSS the least affected (slightly positive), and Qwen3 most affected (negatively, especially on SST-2). Notably, PP is detrimental especially to BO (African American Old) group (especially in Mistral-Medium and Qwen3-32B). This could indicate that PP steers models to have more stereotypical behavior towards this group.
For Qwen3, prompting with persona always decreases performances. 

The results also show that \textbf{models are biased both with and without personas}. All models perform much better on Old age groups, especially on African American and Caucasian (BO, WO), except for LO in SST-2. The same trend was also observed in BRWRR~\cite{thorn-jakobsen-2023-being-right-for-whose-right-reasons}, with 16 transformer models. This may reflect a data imbalance in the many models' pretraining corpora, where the linguistic style of older individuals is better represented or stereotyped as more formal and authoritative.

\paragraph{Rationale Selection} Similar to label prediction, PP did not improve rationale alignment with the corresponding demographic groups over the baselines.
Models exhibit a consistent bias in rationale extraction, showing much higher alignment with the Caucasian (WO) annotator group compared to others. When viewed simultaneously with task performance, models overall have better performance on Caucasian groups.

Between CoS-E and SST-2, PP benefits the former more on rationale selection (although not significantly), and the difference between WO and other groups is more pronounced.

\paragraph{Inter-Persona Agreement}
We calculate the agreements between the 12 personas (Table~\ref{tab:brwrr_alpha_labels_rationales}). 
The agreements based on Krippendorff's $\alpha$ are very high for all models. Only Qwen3 has lower agreements on rationale extraction. This shows that the effect of PP on subtle subjective tasks is limited.

\subsection{Statistical Significance Analysis}

\paragraph{HateXplain} 
As shown in Figure~\ref{fig:hate_labe_baseline_personas}, the impact of persona prompting on label prediction is highly model-dependent. For Mistral-Medium, 11 (out of 21) personas, such as \textit{Atheist}, \textit{Not lonely}, and \textit{White} yield statistically significant improvements over the baseline ($p < 0.05$). 
Conversely, Qwen3-32B exhibits significant degradation across a wide range of personas (16 out of 21), confirming that susceptibility to steering varies by model. 
Crucially, as shown in Figure~\ref{fig:hate_rati_baseline_personas}, these gains in label accuracy do not translate to better rationale selection. We observe no significant improvements in rationale quality for any model-persona pair, and significant degradation in several cases.

\paragraph{BRWRR (CoS-E \& SST-2)}
On these tasks (Figure~\ref{fig:baseline_personas_combined}), the analysis confirms that persona prompting offers no statistically significant benefits.
For GPT-OSS and Mistral-Medium, most differences were statistically insignificant (CI crossing baselines).
For Qwen3-32B, PP led to significant performance drops in both label prediction (e.g., -9.2\% for Black Young Female on SST-2) and rationale selection (Black Old Male).

These results statistically reinforce our finding that PP's utility is task and model-dependent, and often detrimental to rationale quality.

\begin{table}[t]
  \centering
  \large
  \resizebox{\columnwidth}{!}{%
    \begin{tabular}{@{} l r r r r @{}} 
      \hline
      \textbf{Model} & \multicolumn{2}{c}{\textbf{CoS-E}} & \multicolumn{2}{c}{\textbf{SST-2}} \\
      \cline{2-5}
       & \textbf{Label} & \textbf{Rationale} & \textbf{Label} & \textbf{Rationale} \\
      \hline
      GPT-OSS-120B     & $.93 \pm .00$ & $.80 \pm .00$ & $.96 \pm .01$ & $.72 \pm .00$ \\
      Mistral-Medium   & $.95 \pm .00$ & $.85 \pm .00$ & $.98 \pm .00$ & $.89 \pm .01$ \\
      Qwen3-32B        & $.88 \pm .00$ & $.63 \pm .00$ & $.84 \pm .00$ & $.69 \pm .01$ \\
      \hline
    \end{tabular}%
  }
    \caption{Krippendorff's $\alpha$ for labels and rationales on CoS-E and SST-2.}
    \label{tab:brwrr_alpha_labels_rationales}
\end{table}

\begin{table*}[ht]
    \centering
    \begin{adjustbox}{max width=\textwidth}
    \begin{tabular}{l|lc|lc|lc|lc|lc}
    \toprule
    \textbf{Model} & \multicolumn{2}{c}{\textbf{All} \textit{(n=500)}} & \multicolumn{2}{c}{\textbf{Gender} \textit{(n=142)}} & \multicolumn{2}{c}{\textbf{Race/Eth.} \textit{(n=171)}} & \multicolumn{2}{c}{\textbf{Religion} \textit{(n=101)}} & \multicolumn{2}{c}{\textbf{No Target} \textit{(n=294)}} \\
 & Persona & MAE & Persona & MAE & Persona & MAE & Persona & MAE & Persona & MAE \\
\midrule
\multirow{4}{*}{GPT-OSS} & Baseline & 64.3$_{\pm0.1}$ & Baseline & 64.1$_{\pm0.0}$ & Baseline & 54.2$_{\pm1.4}$ & Baseline & 47.2$_{\pm0.5}$ & Baseline & 83.1$_{\pm0.3}$ \\
 & Black & \textbf{56.1}$_{\pm1.8}$ & Black & \textbf{58.0}$_{\pm3.8}$ & Black & \textbf{48.9}$_{\pm3.4}$ & Centrist & \textbf{39.3}$_{\pm0.5}$ & Black & \textbf{70.6}$_{\pm2.3}$ \\
 & Jewish & 60.4$_{\pm1.1}$ & Jewish & 62.2$_{\pm2.6}$ & Hindu & 50.5$_{\pm2.0}$ & Black & 39.9$_{\pm1.7}$ & Jewish & 77.7$_{\pm1.4}$ \\
 & Muslim & 61.1$_{\pm0.9}$ & No Formal Edu. & 62.4$_{\pm2.4}$ & Age 15 & 51.3$_{\pm1.2}$ & Right-wing & 43.6$_{\pm2.9}$ & Muslim & 78.7$_{\pm1.4}$ \\
\midrule
    \multirow{4}{*}{Mistral} & Baseline & 71.7$_{\pm1.2}$ & Baseline & 64.1$_{\pm1.0}$ & Baseline & 59.6$_{\pm0.5}$ & Baseline & 53.5$_{\pm2.1}$ & Baseline & 96.7$_{\pm2.2}$ \\
 & Atheist & \textbf{55.1}$_{\pm0.4}$ & Centrist & \textbf{52.3}$_{\pm1.3}$ & Atheist & \textbf{48.7}$_{\pm0.3}$ & White & \textbf{43.6}$_{\pm1.6}$ & Not Lonely & \textbf{70.9}$_{\pm1.1}$ \\
 & Not Lonely & 55.2$_{\pm0.7}$ & White & 52.6$_{\pm0.7}$ & Not Lonely & 48.9$_{\pm1.2}$ & Not Lonely & 43.9$_{\pm1.7}$ & Atheist & 71.2$_{\pm0.8}$ \\
 & White & 56.4$_{\pm0.6}$ & Atheist & 53.3$_{\pm1.4}$ & White & 50.3$_{\pm1.3}$ & Atheist & 46.2$_{\pm0.5}$ & Centrist & 72.9$_{\pm2.3}$ \\
\midrule
    \multirow{4}{*}{Qwen3} & Baseline & 58.8$_{\pm1.7}$ & Baseline & 56.4$_{\pm2.5}$ & Baseline & 53.8$_{\pm1.3}$ & Baseline & \textbf{44.2}$_{\pm3.7}$ & Baseline & 75.8$_{\pm2.0}$ \\
 & Not Lonely & \textbf{58.7}$_{\pm0.8}$ & Not Lonely & \textbf{54.2}$_{\pm3.5}$ & Left-wing & \textbf{53.2}$_{\pm3.4}$ & Higher Edu. & 49.8$_{\pm2.0}$ & Not Lonely & \textbf{69.4}$_{\pm0.6}$ \\
 & White & 59.5$_{\pm2.3}$ & Atheist & 54.9$_{\pm3.0}$ & Asian & 53.8$_{\pm1.3}$ & Asian & 50.5$_{\pm1.4}$ & Centrist & 69.8$_{\pm0.4}$ \\
 & Atheist & 59.8$_{\pm1.4}$ & Centrist & 55.2$_{\pm2.7}$ & Atheist & 57.7$_{\pm3.6}$ & Male & 51.2$_{\pm0.9}$ & White & 72.6$_{\pm2.9}$ \\
\bottomrule
\end{tabular}   
    \end{adjustbox}
    \caption{Top-3 personas MAE by subgroup (values in \% with mean and std) and model on HateXplain dataset. The top personas of different subgroups are almost consistent with the whole sample set, indicating poor alignment with target groups.} 
    \label{tab:subgroup_f1}
\end{table*}

\section{Discussion}

\paragraph{The effect of PP is task-dependent: label prediction performance gain is not indicative of improved rationale selection.}

Analysis across models and datasets reveals that PP exerts divergent effects on label prediction and rationale selection.
For label prediction, PP improves performance on HateXplain -- especially for Mistral -- but has slightly negative effects on CoS-E and SST-2. For rationale selection, PP produces no gain on any datasets, and mostly worse performances in HateXplain.
Regarding model differences, GPT-OSS and Mistral occasionally outperform their neutral baselines, while Qwen3 consistently deteriorates. 

While HateXplain is the most controversial and socially sensitive topic, CoS-E is also considered subjective (commonsense reasoning) but less socially sensitive, and SST-2 (sentiment analysis on movie reviews) is the least subjective. These differences explain why PP causes more variability in classification performance of HateXplain: our persona attributes are designed to be sensitive to social problems. As CoS-E and SST-2 labels are less ambiguous, PP is less effective.

Looking at the differences in rationale selection performance, we see that PP has the least positive impact on HateXplain. This may indicate that label subjectivity does not mean rationale subjectivity. 
It also suggests that for sensitive topics, model safety alignments create a rigid reasoning template that persona prompts are unable to override. However, our findings on less sensitive tasks reveal that this rigidity is not unique to safety; rather, models exhibit a strong, general resistance to adapting their reasoning style based on persona prompts, echoing the findings of \citet{kim-2024-persona-double-edged-sword}.

\paragraph{Simulated personas are not always aligned with the real-world demographic groups.}
The results in CoS-E and SST-2 show that PP does not always improve performance on the demographic group aligned to that persona, and the difference between personas is minimal (shown by the very high agreement among personas). Furthermore, the best performing persona for a target group is not always from the persona with the same demographic profile (see Table~\ref{tab:best_personas_baseline_cose} and~\ref{tab:best_personas_baseline_sst_f1} in Appendix~\ref{appdix:cose_and_sst}). 

On HateXplain, we do not have the annotators' demographic information, but the ``target'' annotation for each statement. We thus further divide the 500 subsets by different target groups (mainly four subgroups by: \textit{gender}, \textit{ethnicity}, \textit{religion} and \textit{no target}) and check whether the personas improving performance are related to the target subgroups. Shown in Table~\ref{tab:subgroup_f1}, the performance of subgroups where the personas performed the best are not aligned with their target group attributes. Instead, top-3 performing personas for different subgroups are close to the default ranking on the entire data, indicating that personas are not very sensitive to different types of demographics-related issues. 
Notably, we see that the \textit{no target} subgroup performance is much lower than other subgroups because most samples are ``Normal'', and models perform worst on this label due to over-flagging. Additionally, Qwen3 performed worse in Race/Ethnicity compared to Gender and Religion.
Our finding that simulated personas often fail to align with their real-world counterparts urges caution for the growing paradigm of using LLMs as population simulators \cite{argyle2023out}.

\paragraph{Models are biased and strongly guardrailed regardless of PP.}

Our results on all datasets show that bias patterns are present in all models: across CoS-E and SST-2, BO/WO groups outperform BY/WY/LY/LO under all prompting conditions for both labels and rationales. PP does not reduce these biases.
In HateXplain, the positive Mean Error (over-flagging of harmful labels) across personas suggests guardrail-aligned over-flagging by all models. Guardrail sensitivity has been shown to affect classification performance~\cite{rogers2025bias}, our results further indicate this phenomenon.
Inter-persona agreements overall are very high (except Qwen3), suggesting that models possess a reasoning process that is surprisingly resistant to surface-level persona prompts~\cite{liu-2024-evaluating-llm-biases,sommerauer2025simulating}.

\section{Conclusion}
In this paper, we studied how prompting with socio-demographic personas can elicit models' social reasoning by generating word-level rationales. We focused on datasets with different levels of subjectivity: Hate speech detection, commonsense reasoning and sentiment analysis. 
The results across three LLMs reveal three findings: (1) PP shows divergent behavior on task performance and rationales: when predicting labels, models perform better on the more subjective task (hate speech); but when asked to select the rationales, models performed the worse on this task. (2) PP in most cases does not steer models to align better with the corresponding demographic groups. (3) Regardless of personas, models are shown to perform better on specific demographic groups (White, African American, Old) and heavily over-flag harmful content. 
We believe these findings highlight a fundamental challenge for personalized AI: true alignment likely requires more than persona prompting, and future work must develop methods that can genuinely steer a model's underlying reasoning process without sacrificing performance or reinforcing existing~biases.

\section*{Limitations}

This study is bounded by certain limitations, starting with the datasets' ground truths. Each sample in HateXplain is annotated by only three individuals, with disagreements resolved through majority vote, which may not capture the full spectrum of interpretations. Secondly, our rationales are generated by model prompting, instead of via traditional feature attribution methods. Thus, the rationales may not be faithful to a model's internal activations.
Finally, we selected three representative models in the study, and Qwen3 model diverges the most from the other two models, indicating that other LLMs may show different behaviors than what we have observed.

\section*{Ethical Statement}
Our research aims to contributing to safer and more fair models. However, challenges such as incorrect classifications can impact the reliability of these systems and potentially affect users negatively. Additionally, model outputs can vary substantially depending on the persona simulated or the particular language model employed. Given this variability, we caution against placing excessive trust in automated detection tools, as it may reduce users' careful judgment.

\section*{Acknowledgments}
We thank the anonymous reviewers at ACL Rolling Review for their valuable feedback.
This research is funded by the Berlin Institute for the Foundations of Learning and Data (BIFOLD, ref. 01IS18037A) and the Federal Ministry of Education and Research (BMFTR) in the scope of the research project VeraXtract (ref. 16IS24066).

\section*{Contributions}
JY and NF conceptualized the project and jointly supervised MH, EK, and ELB, who provided the initial implementation and writing in the scope of a course project at TU Berlin.
JY and MH implemented and ran further experiments.
JY and NF conducted further analyses of the results.
JY, MH, VS, and NF contributed to the final draft.

\bibliography{custom}

\clearpage
\appendix

\section{Experimental details}
\label{appdix:exp_detail}

\subsection{Single-persona attribute selection}
\label{appdix:persona_hate}

Table~\ref{tab:single_attribute_personas} shows the values for the seven chosen demographic attributes. 

\begin{table}[ht]
\centering
\small
\renewcommand{\arraystretch}{1.1}
\begin{tabularx}{0.85\linewidth}{@{}l X@{}}
\toprule
\textbf{Attributes} & \textbf{Values} \\
\midrule
Age & 15, 35, 65 \\
\midrule
Education & No formal education, High school education, Higher education \\
\midrule
Gender & Male, Female \\
\midrule
Loneliness & Not lonely, Somewhat lonely \\
\midrule
Political view & Left-wing, Right-wing, Centrist \\
\midrule
Race & White, Black, Asian \\
\midrule
Religion & Christian, Muslim, Jewish, Atheist, Hindu \\
\bottomrule
\end{tabularx}
\caption{Socio-demographic categories and values.}
\label{tab:single_attribute_personas}
\end{table}

We list below our reasons for choosing the attributes to construct our personas in HateXplain. 

\begin{itemize}
    \item Loneliness: Due to the COVID-19 pandemic and digitalisation of our society, the lack of social interactions has a great impact on us. Lonely individuals often experience more negative emotions and therefore are more likely to interpret neutral or ambiguous statements as hostile. Because of the lack of social interaction they may lack ``social calibration'' and struggle more with judging whats appropriate or not. And they might have heightened sensitivity for rejection because of self protection.
    \item Age, gender and race: These are classic demographic attributes. They provide a foundational way of grouping people and are often the first factors considered in social science research.
    \item Education: Education is closely tied to intellectual development. We expect that differences in academic exposure or general knowledge levels may influence how people perceive or engage with potentially harmful speech.
    \item Political view and religion: They represent areas with high potential for disagreement and controversy. These categories often shape people’s values, worldviews, and sensitivities, which makes them particularly relevant when analyzing attitudes toward hate speech.
\end{itemize}

\subsection{Implementation details}
\label{appdix:imp_detail}
For all models, we use their default sampling parameters. For GPT-OSS-120B, we set thinking effort to be \textbf{medium}, as we consider the tasks to be relatively simple. We perform one prompt per persona per instance, resulting in 31,500 completions (500 instances × 21 personas × 3 runs) for HateXplain, 18,000 completions (500 instances × 12 personas × 3 runs) for CoS-E, and 9,468 completions (263 instances × 12 personas × 3 runs) for SST-2, per model, in addition to the baseline generation (also 3 runs per model). For all completions, we use an API key from OpenRouter~\footnote{\url{https://openrouter.ai/}}.

Regarding computational costs, we ended up with $\approx\$32$ for GPT-OSS-120B + $\approx\$5$ for Mistral-Medium (we initially used Mistral Experimental API which was free, but we hit rate limits later on and used the paid API access) + $\approx\$30$ for Qwen3-32B. 
With respect to inference time, the Mistral model was fastest as it is not a reasoning model (each run takes around 2h), and GPT-OSS-120B had also faster response as it is a MoE model, while Qwen3-32B had the slowest response.

\clearpage
\section{Prompts}
\label{appdix:prompts}

All prompts are used in user prompt. We attempted putting persona related content in system prompt but the reasonings show that with user prompt, models act better as that persona.
\subsection{Baseline Prompts}
\begin{promptbox}[HateXplain Baseline Prompt]
Given the following text post, classify it into one of the available categories:

Available categories: Hate speech, Offensive language, Normal

First, think step by step, explaining your reasoning inside \texttt{<think>...</think>} tags.

Then, respond with a JSON object in this format (and no other text):

\begin{codechunk}
{
  "label": "<one of the above categories>",
  "rationale": ["word1", "word2", "..."],
  "reasoning": "<your reasoning>" # only for Mistral-Medium
}
\end{codechunk}

Input:
\begin{verbatim}
<input_text_here>
\end{verbatim}
\end{promptbox}


\begin{promptbox}[CoS-E Baseline Prompt]
Task: first select the answer you think is most correct and sensible; then mark relevant words in the question that justify your choice.

First, think step by step inside \texttt{<think>...</think>} tags.

Then, respond only with a JSON object in this format (and no other text):

\begin{codechunk}
{
  "answer": "<exact option text>",
  "answer_index": <0-based index>,
  "rationale": ["keyword1", "keyword2"],
  "reasoning": "<your reasoning>" # only for Mistral-Medium
}
\end{codechunk}

Question:
\begin{verbatim}
<question_text_here>
\end{verbatim}

Options:
\begin{codechunk}
0. <option 0>
1. <option 1>
2. <option 2>
...
\end{codechunk}

Rules: \\
- The value for "answer" must be copied verbatim from the options list.\\
- "answer\_index" must match the position (0-based) in the options list.\\
- For each word in the question, if removing it would lower your confidence in the chosen label, include it in "rationale".
\end{promptbox}

\begin{promptbox}[SST-2 Baseline Prompt]
Task: read the following sentence and choose exactly one of three labels.

Positive: The sentence conveys information about the author’s positive evaluative sentiment.

Negative: The sentence conveys information about the author’s negative evaluative sentiment.

No sentiment: The sentence does not convey anything about the author’s positive or negative sentiment.

Please specify what snippets of text, in the sentence, act as supporting evidence for your chosen label.

First, think step by step inside \texttt{<think>...</think>} tags.

Then, respond only with a JSON object in this format (and no other text):

\begin{codechunk}
{
  "label": "<one of the above categories>",
  "rationale": ["word1", "word2"],
  "reasoning": "<your reasoning>"
}
\end{codechunk}

Sentence:
\begin{verbatim}
<sentence_here>
\end{verbatim}
\end{promptbox}

\subsection{Persona Prompts}
\begin{promptbox}[HateXplain Persona Prompt]
Step into the shoes of a real person who \texttt{<persona description>}.

Imagine you have lived your whole life shaped by this trait. How might it affect your thoughts, emotions, sensitivities, and judgments?

Given the following text post, analyze and classify it exactly as this person would—letting their background, beliefs, and life experiences guide your reasoning.

Available categories for classification:
Hate speech, Offensive language, Normal

First, \emph{think out loud} as this persona, step by step, inside \texttt{<think>...</think>} tags. Reflect honestly about how this person's perspective influences what they notice and how they judge the text.

Then, respond only with a JSON object in this format (and no other text):

\begin{codechunk}
{
  "label": "<one of the above categories>",
  "rationale": ["word1", "word2", "..."],
  "reasoning": "<your reasoning>" # only for Mistral-Medium
}
\end{codechunk}

- Stay fully in character; do not break persona or add explanations outside the JSON.
\\ \\
Input:
\begin{verbatim}
<input_text_here>
\end{verbatim}
\end{promptbox}

\begin{promptbox}[CoS-E Persona Prompt]
Step into the shoes of a real person who \texttt{<persona description>}.

Imagine you have lived your whole life shaped by this trait. How might it affect your thoughts, emotions, sensitivities, and judgments?

Task: first select the answer you think is most correct and sensible; then mark relevant words in the question that justify your choice.

First, \emph{think out loud} as this persona, step by step, inside \texttt{<think>...</think>} tags. Reflect honestly about how this person's perspective influences what they notice and how they judge the text.

Answer it exactly as this person would—letting their background, beliefs, and life experiences guide your reasoning.

Then, respond only with a JSON object in this format (and no other text):

\begin{codechunk}
{
  "answer": "<exact option text>",
  "answer_index": <0-based index>,
  "rationale": ["keyword1", "keyword2"],
  "reasoning": "<your reasoning>"
}
\end{codechunk}

Question:
\begin{verbatim}
<question_text_here>
\end{verbatim}

Options:
\begin{codechunk}
0. <option 0>
1. <option 1>
2. <option 2>
...
\end{codechunk}

Rules:

- The value for "answer" must be copied verbatim from the options list.\\
- "answer\_index" must match the position (0-based) in the options list.\\
- For each word in the question, if removing it would lower your confidence in the chosen label, include it in "rationale".\\
- Stay fully in character; do not break persona or add explanations outside the JSON.
\end{promptbox}

\begin{promptbox}[SST-2 Persona Prompt]
Step into the shoes of a real person who is \texttt{<persona description>}.

Imagine you have lived your whole life shaped by this trait. How might it affect your thoughts, emotions, sensitivities, and judgments?

Task: read the following sentence and choose from one of the three labels:

Positive: The sentence conveys information about the author’s positive evaluative sentiment.

Negative: The sentence conveys information about the author’s negative evaluative sentiment.

No sentiment: The sentence does not convey anything about the author’s positive or negative sentiment.

Please specify what snippets of text, in the sentence, you think act as supporting evidence for your chosen label.

First, \emph{think out loud} as this persona, step by step, inside \texttt{<think>...</think>} tags. Reflect honestly about how this person's perspective influences what they notice and how they judge the text.

Answer it exactly as this person would—letting their background, beliefs, and life experiences guide your reasoning.

Then, respond only with a JSON object in this format (and no other text):

\begin{codechunk}
{
  "label": "<one of the above categories>",
  "rationale": ["word1", "word2"],
  "reasoning": "<your reasoning>"
}
\end{codechunk}

Sentence:
\begin{verbatim}
<sentence_here>
\end{verbatim}

Rules:

- Stay fully in character; do not break persona or add explanations outside the JSON.
\end{promptbox}

\clearpage
\section{Additional results}

\subsection{CoT vs. no-CoT Prompting Comparison}
\label{appendix:cot_no_cot}

We compared the results with and without the think-step-by-step reasoning instruction (CoT) and show them in Table~\ref{tab:hatexplain_results}, \ref{tab:cose_results} and \ref{tab:sst2_results}. In general, without CoT prompt, models perform slightly worse than with CoT prompt. The largest difference is in rationale selection (token-$F_1$) with the Mistral model; as a non-reasoning model, it benefits more from CoT in terms of rationale selection.

\begin{table*}[ht]
\centering
\resizebox{\textwidth}{!}{%
\begin{tabular}{lcccc|cccc}
\toprule
\multirow{3}{*}{\textbf{Model}} & \multicolumn{4}{c}{\textbf{Label Performance (Macro-$F_1$)}} & \multicolumn{4}{c}{\textbf{Rationale Performance (Token-$F_1$)}} \\
\cmidrule(lr){2-5} \cmidrule(lr){6-9}
 & \multicolumn{2}{c}{Baseline} & \multicolumn{2}{c|}{Persona Average} & \multicolumn{2}{c}{Baseline} & \multicolumn{2}{c}{Persona Average} \\
 & CoT & No-CoT & CoT & No-CoT & CoT & No-CoT & CoT & No-CoT \\
\midrule
GPT-OSS-120B & $48.8_{\pm 0.2}$ & $48.2_{\pm 0.7}$ & $48.4_{\pm 1.3}$ & $47.8_{\pm 1.8}$ & $69.6_{\pm 1.8}$ & $68.5_{\pm 0.4}$ & $69.2_{\pm 0.9}$ & $67.0_{\pm 2.3}$ \\
Mistral-Medium & $45.6_{\pm 0.8}$ & $44.3_{\pm 0.7}$ & $45.9_{\pm 3.9}$ & $47.2_{\pm 3.4}$ & $66.1_{\pm 0.2}$ & $67.6_{\pm 0.6}$ & $64.6_{\pm 1.2}$ & $60.3_{\pm 1.7}$ \\
Qwen3-32B & $50.9_{\pm 0.6}$ & $48.8_{\pm 1.6}$ & $43.8_{\pm 3.0}$ & $42.8_{\pm 3.0}$ & $67.7_{\pm 0.5}$ & $66.5_{\pm 2.1}$ & $66.2_{\pm 1.2}$ & $65.6_{\pm 1.3}$ \\
\bottomrule
\end{tabular}%
}
\caption{HateXplain performance comparison of Baseline and Persona settings across CoT and No-CoT prompting strategies. Values are reported in percentages as $Mean_{\pm SD}$.}
\label{tab:hatexplain_results}
\end{table*}

\begin{table*}[h!]
\centering
\resizebox{\textwidth}{!}{%
\begin{tabular}{lcccc|cccc}
\toprule
\multirow{3}{*}{\textbf{Model}} & \multicolumn{4}{c}{\textbf{Label Performance (Accuracy \%)}} & \multicolumn{4}{c}{\textbf{Rationale Performance (Token-$F_1$)}} \\
\cmidrule(lr){2-5} \cmidrule(lr){6-9}
 & \multicolumn{2}{c}{Baseline} & \multicolumn{2}{c|}{Persona Average} & \multicolumn{2}{c}{Baseline} & \multicolumn{2}{c}{Persona Average} \\
 & CoT & No-CoT & CoT & No-CoT & CoT & No-CoT & CoT & No-CoT \\
\midrule
GPT-OSS-120B & $70.5_{\pm 3.7}$ & $71.0_{\pm 3.8}$ & $70.4_{\pm 3.4}$ & $70.6_{\pm 3.4}$ & $53.8_{\pm 3.0}$ & $53.3_{\pm 3.1}$ & $54.1_{\pm 2.8}$ & $53.7_{\pm 2.9}$ \\
Mistral-Medium & $71.0_{\pm 3.8}$ & $70.6_{\pm 3.2}$ & $70.3_{\pm 2.9}$ & $69.9_{\pm 2.9}$ & $52.0_{\pm 3.3}$ & $49.8_{\pm 2.9}$ & $52.8_{\pm 3.1}$ & $51.7_{\pm 2.9}$ \\
Qwen3-32B & $70.4_{\pm 4.1}$ & $70.5_{\pm 4.1}$ & $69.1_{\pm 3.4}$ & $68.7_{\pm 3.3}$ & $49.3_{\pm 3.2}$ & $48.9_{\pm 3.0}$ & $48.0_{\pm 2.7}$ & $47.8_{\pm 2.9}$ \\
\bottomrule
\end{tabular}%
}
\caption{CoS-E performance comparison of Baseline and Persona settings across CoT and No-CoT prompting strategies. Values are reported in percentages as $Mean_{\pm SD}$.}
\label{tab:cose_results}
\end{table*}

\begin{table*}[h!]
\centering
\resizebox{\textwidth}{!}{%
\begin{tabular}{lcccc|cccc}
\toprule
\multirow{3}{*}{\textbf{Model}} & \multicolumn{4}{c}{\textbf{Label Performance (Binary $F_1$)}} & \multicolumn{4}{c}{\textbf{Rationale Performance (Token-$F_1$)}} \\
\cmidrule(lr){2-5} \cmidrule(lr){6-9}
 & \multicolumn{2}{c}{Baseline} & \multicolumn{2}{c|}{Persona Average} & \multicolumn{2}{c}{Baseline} & \multicolumn{2}{c}{Persona Average} \\
 & CoT & No-CoT & CoT & No-CoT & CoT & No-CoT & CoT & No-CoT \\
\midrule
GPT-OSS-120B & $95.7_{\pm 2.0}$ & $95.7_{\pm 1.9}$ & $96.1_{\pm 1.9}$ & $96.0_{\pm 2.0}$ & $58.4_{\pm 4.4}$ & $58.7_{\pm 4.3}$ & $57.4_{\pm 3.6}$ & $57.4_{\pm 3.6}$ \\
Mistral-Medium & $96.0_{\pm 2.2}$ & $96.0_{\pm 2.2}$ & $95.2_{\pm 1.9}$ & $95.3_{\pm 1.9}$ & $57.5_{\pm 4.2}$ & $56.1_{\pm 4.3}$ & $56.9_{\pm 3.6}$ & $56.5_{\pm 3.7}$ \\
Qwen3-32B & $95.4_{\pm 1.9}$ & $95.8_{\pm 2.0}$ & $92.1_{\pm 3.0}$ & $91.6_{\pm 2.8}$ & $58.1_{\pm 4.5}$ & $57.9_{\pm 4.2}$ & $56.0_{\pm 3.7}$ & $54.1_{\pm 3.6}$ \\
\bottomrule
\end{tabular}%
}
\caption{SST-2 performance comparison of Baseline and Persona settings across CoT and No-CoT prompting strategies. Values are reported in percentages as $Mean_{\pm SD}$.}
\label{tab:sst2_results}
\end{table*}

\subsection{HateXplain Additional Results}
\label{appdix:hate}
While inter-annotator agreement metrics like Krippendorff's $\alpha$ suggest general consistency within demographic aspects, they do not reveal systematic shifts in label distribution between specific personas. 
To investigate the impact of each demographic aspect, we performed pairwise Stuart-Maxwell tests on the classification distributions of personas within each category (Table~\ref{tab:stuart_maxwell_summary}). We applied a Bonferroni correction to strictly account for multiple comparisons.

The results confirm that despite high agreement, significant distributional shifts exist. Political View and Loneliness showed significant differences across almost all models. For instance, in the Political View category (Table~\ref{tab:political_distribution_tests}), Left-wing personas consistently predicted ``Hate Speech'' at a significantly higher rate (e.g., 53\% for Mistral-Medium) compared to Right-wing (32\%) and Centrist (24\%) personas. Conversely, Right-wing personas were more likely to classify content as ``Normal'' or ``Offensive''.

To quantify the magnitude of these differences intuitively, we calculated the percentage of samples where personas within the same aspect disagreed (Table~\ref{tab:disagreement_percentages}). Political View again emerged as a high-disagreement category (up to 56.4\% for Qwen3-32B), aligning with our statistical findings. Notably, Qwen3-32B exhibited substantially higher disagreement rates across all attributes compared to GPT-OSS-120B and Mistral-Medium, suggesting it is less stable under persona prompting.

\begin{figure*}[!ht]
    \centering
    \includegraphics[width=\linewidth]{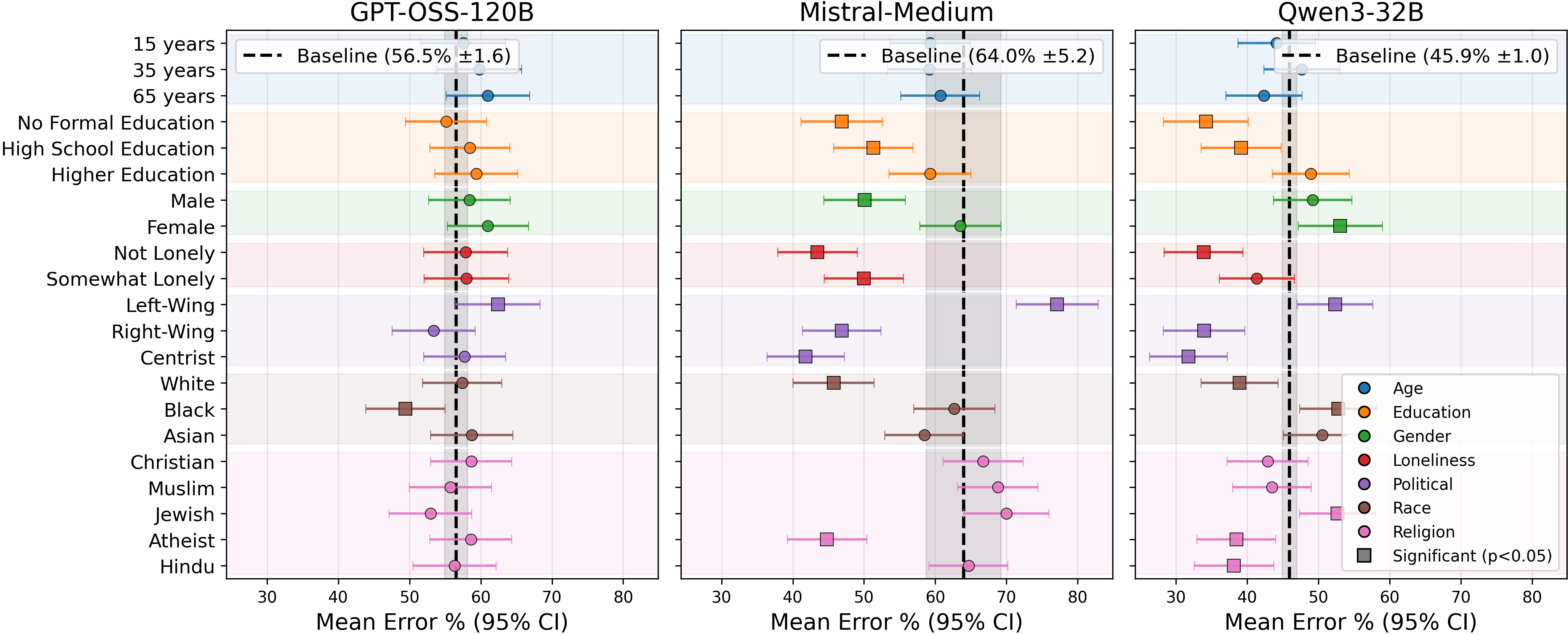}
    \caption{Mean Error (ME) of baseline and personas in label prediction on HateXplain dataset. Error bars incidates 95\% CIs. ME of GPT-OSS and Mistral are similar to MAE of them, while ME of Qwen3 showing some personas have lower ME, but higher MAE than baseline. Lower ME indicates generally lower over-flagging rates.}
\label{fig:hate_me_baseline_personas}
\end{figure*}

\begin{figure*}[!ht]
    \centering
    \includegraphics[width=\linewidth]{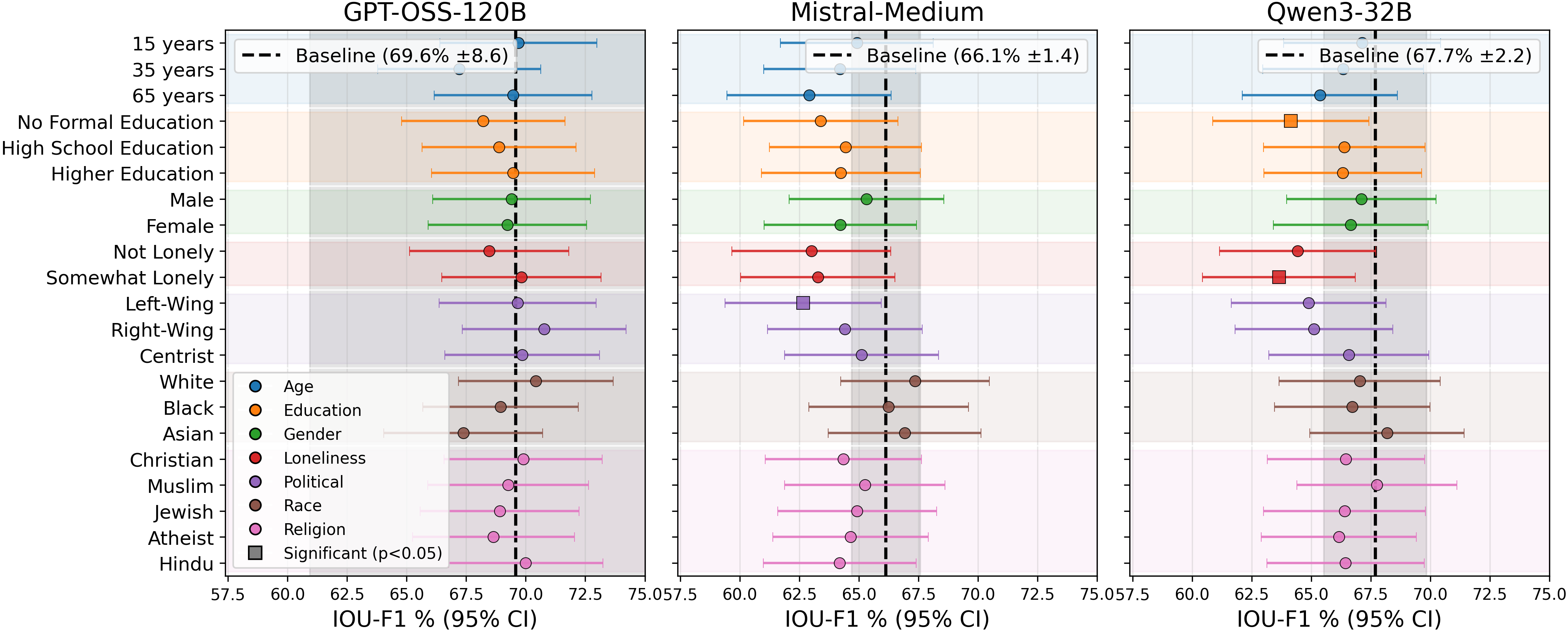}
    \caption{Rationale IOU-$F_1$ of baseline (no persona) and different single-attribute persona performances on HateXplain dataset, excluding rationales from ``normal'' label. Error bars incidates 95\% CIs.}
    \label{fig:hate_iou_baseline_personas}
\end{figure*}

We additionally show the results of ME and IOU-$F_1$ performance in Figure~\ref{fig:hate_me_baseline_personas} and~\ref{fig:hate_iou_baseline_personas}. An additional CoT reasoning example for \textit{No formal education} persona is shown in Table~\ref{tab:cot_consistency}.

\begin{table}[H]
    \centering
    \Large
    \resizebox{\columnwidth}{!}{%
        \begin{tabular}{l c c c c c}
        \toprule
        \textbf{Attribute} & \textbf{\# Tests} & \textbf{Bonf. $\alpha$} & \textbf{GPT-OSS} & \textbf{Mistral} & \textbf{Qwen3} \\
        \midrule
        Loneliness      & 1  & 0.0500 & 1/1  & 1/1  & 1/1  \\
        Political View  & 3  & 0.0167 & 2/3  & 3/3  & 3/3  \\
        Race            & 3  & 0.0167 & 2/3  & 3/3  & 2/3  \\
        Religion        & 10 & 0.0050 & 2/10 & 8/10 & 7/10 \\
        Age             & 3  & 0.0167 & 1/3  & 2/3  & 2/3  \\
        Education       & 3  & 0.0167 & 0/3  & 3/3  & 2/3  \\
        Gender          & 1  & 0.0500 & 0/1  & 1/1  & 0/1  \\
        \bottomrule
    \end{tabular}%
    }
    \caption{Stuart-Maxwell Test results comparing label distributions of personas within each demographic aspect. Each cell indicates the number of significant pairwise comparisons out of the total number of tests performed ($k$). A Bonferroni correction was applied to the significance level $\alpha$ based on $k$.}
    \label{tab:stuart_maxwell_summary}
\end{table}

\begin{table*}[ht!]
    \centering
    \resizebox{\textwidth}{!}{%
    \begin{tabular}{l l c l l}
        \toprule
        \textbf{Model} & \textbf{Comparison} & \textbf{Sig.} & \textbf{Persona 1 Distribution} & \textbf{Persona 2 Distribution} \\
        \midrule
        GPT-OSS-120B   & Left-wing vs. Right-wing & $\checkmark^{***}$ & 52\% HS, 22\% Off, 26\% Norm & 46\% HS, 25\% Off, 29\% Norm \\
                       & Left-wing vs. Centrist   & $\checkmark^{***}$ & 52\% HS, 22\% Off, 26\% Norm & 47\% HS, 28\% Off, 25\% Norm \\
                       & Right-wing vs. Centrist  & $\times^{*}$       & 46\% HS, 25\% Off, 29\% Norm & 47\% HS, 28\% Off, 25\% Norm \\
        \midrule
        Mistral-Medium & Left-wing vs. Right-wing & $\checkmark^{***}$ & 53\% HS, 35\% Off, 12\% Norm & 32\% HS, 46\% Off, 21\% Norm \\
                       & Left-wing vs. Centrist   & $\checkmark^{***}$ & 53\% HS, 35\% Off, 12\% Norm & 24\% HS, 57\% Off, 18\% Norm \\
                       & Right-wing vs. Centrist  & $\checkmark^{***}$ & 32\% HS, 46\% Off, 21\% Norm & 24\% HS, 57\% Off, 18\% Norm \\
        \midrule
        Qwen3-32B      & Left-wing vs. Right-wing & $\checkmark^{***}$ & 30\% HS, 56\% Off, 13\% Norm & 24\% HS, 51\% Off, 25\% Norm \\
                       & Left-wing vs. Centrist   & $\checkmark^{***}$ & 30\% HS, 56\% Off, 13\% Norm & 16\% HS, 63\% Off, 21\% Norm \\
                       & Right-wing vs. Centrist  & $\checkmark^{***}$ & 24\% HS, 51\% Off, 25\% Norm & 16\% HS, 63\% Off, 21\% Norm \\
        \bottomrule
    \end{tabular}%
    }
    \caption{Pairwise comparison of label distributions for Political View personas. \textbf{Sig.} denotes statistical significance under the Stuart-Maxwell test with Bonferroni correction. HS: Hate Speech, Off: Offensive, Norm: Normal. Left-wing personas consistently predict Hate Speech at a higher rate than Right-wing or Centrist personas.}
    \label{tab:political_distribution_tests}
\end{table*}

\begin{table}[H]
    \centering
    \small
    \resizebox{\columnwidth}{!}{%
    \begin{tabular}{l c c c}
        \toprule
        \textbf{Attribute} & \textbf{GPT-OSS} & \textbf{Mistral} & \textbf{Qwen3} \\
        \midrule
        Age             & 17.6\% & 17.8\% & 45.6\% \\
        Loneliness      & 14.4\% & 13.4\% & 36.8\% \\
        Religion        & 29.8\% & 36.4\% & 59.2\% \\
        Education       & 22.6\% & 23.8\% & 51.4\% \\
        Race            & 23.8\% & 26.8\% & 45.8\% \\
        Gender          & 14.8\% & 16.6\% & 28.0\% \\
        Political View  & 26.2\% & 49.6\% & 56.4\% \\
        \bottomrule
    \end{tabular}%
    }
    \caption{Percentage of samples where personas within the same demographic aspect disagreed on the predicted label. Qwen3-32B exhibits the highest disagreement rates across all attributes, while Political View and Religion generate the most divergence across all models.}
    \label{tab:disagreement_percentages}
\end{table}

\begin{table*}[ht]
  \centering
  \resizebox{\textwidth}{!}{%
    \begin{tabular}{p{1.6cm}p{24cm}}
      \toprule
        \textbf{Model} & \textbf{Chain-of-Thought reasoning} \\

        GPT OSS, Run 1 
            & \footnotesize 
            The task: We must adopt perspective of a real person with no formal education, whose entire life shaped by this trait. We need to think out loud as that person about the post, then output JSON with label and rationale: words that contributed. We must classify the text as Hate speech, Offensive language, or Normal. The text: "antifa are the real nazis man". This is a statement that accuses antifa of being Nazis. Is it hateful? It's a political statement, possibly offensive. Let's see classification guidelines: Typically hate speech is targeting protected group based on race, religion, etc. "antifa" is political ideology, not protected class, but could be hate speech if inciting violence? It's a derogatory statement possibly offensive but not hate speech. Offensive language? It's an insult but not profanity. It's an expression of political opinion; could be considered normal speech but possibly offensive to antifa supporters. Likely classify as Offensive language? The categories: Hate speech, Offensive language, Normal. I'd choose Offensive language because it's insulting and could be considered offensive, though not hate speech. So label Offensive language. Rationale words: "antifa", "nazis". Those are key. Now we must think out loud as persona with no formal education. They might think of it as rude. We'll produce think tags. Need to output JSON only. So final.Alright, read that line: "antifa are the real nazis man". It's sayin' that a group called antifa are like Nazis. I know a bit about the world, folks that call someone a Nazi is usually meanin' they think they're evil or bigoted. It's a harsh thing to say about a group. It's not a polite thing; it's an insult. There's no fancy words or school talk, just plain talk. It sounds offensive, kinda wantin' to stir up feelings against antifa. It's not exactly talk about race or religion, but it is a strong insult. So I think this is offensive language, not just normal talk. The words that make it stand out are "antifa" and "nazis".
            \textbf{Offensive language}
            \\
        GPT OSS, Run 2
            & \footnotesize 
            We need to produce a response with thinking in <think> tags in persona of a real person with no formal education. Then final JSON. Must stay in character, not break. So we need to simulate inner monologue of person no formal education, describing how they think, then classify text as Hate speech etc. The text: "antifa are the real nazis man". It's a statement accusing antifa of being Nazis. That is offensive maybe hateful towards a group (Antifa). It's a derogatory statement and likely hate speech? In many moderation guidelines, "antifa are the real nazis" is extremist hate speech targeting a political group. It's hateful content. So label "Hate speech". Provide rationale array with words that triggered, e.g., "antifa", "nazis". Also "real". Possibly "man"? Only include words that contributed. We'll produce JSON with label "Hate speech" and rationale list. Also include think block before JSON. We must ensure the think block is inside <think>...</think> tags. Then after that output JSON only. No extra text. So final answer: <think>... </think> {  "label": "Hate speech",  "rationale": ["antifa", "nazis"] } Make sure to not include extra spaces.I ain't got no school books, but I know folks can get hurt by words. When I hear someone say "antifa are the real nazis man", it feels like they're telling a whole bunch of people they’re like the worst kind of hateful group. Calling a group "Nazis" is a strong insult, and it's pointing at a political group, which can stir up hate. In my world, I know that kind of talk can make people angry or scared, and it’s more than just a simple opinion—it’s a hateful label. So I'd put this in the hate speech bucket.
            \textbf{Hate speech}
            \\
        GPT OSS, Run 3
            & \footnotesize 
            We need to adopt persona of a real person with no formal education, shaped by this trait all their life. The classification categories: Hate speech, Offensive language, Normal. The text "antifa are the real nazis man". The user wants the assistant to think out loud as this persona, then output JSON with label and rationale (list of words). The persona: no formal education, maybe limited vocabulary, possibly working class, possibly influenced by certain beliefs, perhaps anti-establishment or strong opinions, but lacking formal education. We need to imagine how they'd think about the text. They might see that calling antifa "real nazis" is an insult, could be hateful towards a political group, but maybe they don't see it as hate speech; maybe they'd think it's just an opinion, a strong statement. But we must decide classification based on typical content guidelines: "antifa are the real nazis man" is a statement that equates a political group with Nazis. That could be considered hate speech? According to policies, hate speech is content that attacks a protected group based on protected characteristics (race, religion, etc.). However, antifa is a political ideology, not a protected characteristic. So it's not hate speech per policy. Offensive language? The phrase includes "real nazis" which is offensive but not profanity. It's a slur-like accusation but not profanity. Could be considered harassment because calling a group Nazis is harassing? Usually hate speech policy: targeted harassment of a protected group. Antifa is not a protected group. So likely classification is "Normal". The rationale list should include the words that contributed: maybe "antifa", "nazis". Possibly "real". We need to output JSON with label and rationale array of words. Now we need to think in persona. Provide <think>...</think> with interior monologue, reflecting the perspective of a person with no formal education. Then output JSON. Make sure output includes only JSON after the think block. No extra text. Let's craft think: <think>...some reasoning ...</think> Then JSON: {"label":"Normal","rationale":["antifa","nazis"]} Yes.I ain't got fancy school, but I know people can get all riled up talkin' about groups. This here says "antifa are the real nazis man". To me, it's just someone shoutin' an opinion, callin' a group named Antifa the same as Nazis. It's a harsh thing, but it's not swearin' or a slur against a race or religion. It's more political talk. I wouldn't call it hate speech 'cause it's not about a protected group. It's kinda offensive maybe, but it's just talk. So I reckon it's normal kind of speech, just loud.
            \textbf{Normal}
            \\
      \midrule
        Mistral, Run 1
            & \footnotesize 
            I don't know much about big words or politics, but it seems like the person is just saying what they think. They ain't cussing or calling names in a mean way, just saying their piece. Don't see nothing too bad here, just words I've heard before.
            \textbf{Normal}
            \\
        Mistral, Run 2
            & \footnotesize 
            I don't know much about big words or politics, but it sounds like someone just saying what they think. Ain't no cussing or calling names in a mean way, just talking about groups they don't like. Folks say stuff like this all the time where I'm from, and it don't seem like it's trying to hurt nobody directly.
            \textbf{Normal}
            \\
        Mistral, Run 3
            & \footnotesize 
            I don't know much about big words or politics, but it sounds like the person is just saying what they think. They ain't cussing or calling names in a mean way, just talking about groups. It don't seem hateful or offensive to me, just regular talk.
            \textbf{Normal}
            \\
      \midrule
        Qwen3, Run 1  
            & \footnotesize 
            Okay, so I need to figure out how someone without formal education would classify the statement "antifa are the real nazis man." Let me start by thinking about their perspective. First, this person probably doesn't have a lot of knowledge about political groups or historical terms. They might not know what Antifa or Nazis really are. They rely on what they hear from people around them or media they consume, maybe from sources that aren't very educational. The statement is making a comparison between Antifa and Nazis. Without formal education, they might not understand that Antifa is a movement against fascism, while Nazis are far-right, racist, and oppressive. If they don't know the actual meanings, they might just take the words at face value. They might have heard "Antifa" in a negative context, maybe from someone they trust or from media that portrays them negatively. So they might believe that calling Antifa "the real Nazis" is just repeating what they've heard. Since they don't have the background to know better, they might not see the inaccuracy. Now, offensive language usually includes slurs or harsh terms. "Nazis" is a strong word, but in this case, the person is using it as a label for Antifa. Since they don't understand the historical context, they might not realize the offensive nature of the term. They could think it's just a strong way to criticize Antifa. Hate speech involves targeting a group with intent to harm or degrade. But without understanding the groups involved, the person might not recognize that Antifa and Nazis are on opposite ends. They might not see the contradiction, so they might not classify it as hate speech. So, putting this all together: the person doesn't know the actual meanings, so they might see it as a normal statement because they're just repeating what they've heard. The words "nazi" and "antifa" are there, but they don't understand the context. So the classification would be "Normal" because from their perspective, it's just an opinion they don't find offensive or hateful.
            \textbf{Normal}
            \\
        Qwen3, Run 2
            & \footnotesize
            Okay, let's try to figure out how someone with no formal education might classify the statement "antifa are the real nazis man." First, I need to imagine what their life experiences are like. Without formal education, this person might rely more on what they hear in everyday conversations, maybe from people they trust or in their community. They might not be familiar with historical context or political terms, so they could be using words they heard without understanding their actual meanings. The phrase "antifa are the real nazis" is a common far-right talking point. Someone without formal education might not know what "antifa" stands for. They might have heard it's a group that opposes something, maybe even violence, but they're using the term "nazi" as an insult, which they think is a bad label. They might not understand the historical significance of the word "nazi," so they might not realize it's offensive in a different context. Looking at the words used, "nazi" is a strong term. But if they don't grasp its meaning, they might not see it as offensive. They might just be using it as a bad word they heard others use. The person might not recognize that calling antifa (which is short for anti-fascist) a nazi is a contradiction, since antifa is against fascism, which naziism is a part of. Without that knowledge, they might not see the irony or the incorrect use of the term. So, from their perspective, they might not think the statement is offensive. They might see it as a normal opinion, even if it's based on a misunderstanding. They might not have the vocabulary or context to label it as hate speech or offensive language. Therefore, they might classify it as "Normal" because they don't understand the deeper implications or the misuse of the term "nazi."
            \textbf{Normal}
            \\
        Qwen3, Run 3  
            & \footnotesize 
            Okay, so I need to figure out how a person without formal education would classify the statement "antifa are the real nazis man." Let me start by thinking about how someone with no formal education might perceive this. First, they might not have a deep understanding of political terms like "antifa" or "nazis." Antifa usually refers to anti-fascist groups, and Nazis are associated with extreme right-wing, racist ideologies. But without formal education, the person might not know the historical context of the Nazis. They might have heard the word "nazi" used colloquially to describe someone they disagree with politically, maybe from people around them or media they consume. So the statement is comparing antifa to Nazis. The person might not know that antifa is actually against fascism, which the Nazis represent. They might just take the words at face value. If they've been taught or have been exposed to the idea that certain groups are bad (like Nazis), and someone is calling another group (antifa) by that name, they might see it as a negative label.  They might not recognize the irony or the incorrect comparison. Since they don't have the education to understand the historical context, they might just see it as someone using a bad word to insult another group. In that case, they might classify this as hate speech because it's using a derogatory term (nazi) to attack a group (antifa).  But wait, offensive language is different from hate speech. Hate speech is more about targeting a group based on race, religion, etc. Here, the term "nazi" is being used as an insult. If the person doesn't know the actual meaning, they might not see it as hate speech, but just as offensive language. However, if they associate "nazi" with extreme negativity, they might label it as hate speech because they think it's a serious accusation against a group. Also, the word "man" at the end is just an exclamation, so it's not key. The key words are "antifa" and "nazi." Since the person might not know what antifa stands for, they might just take the statement as a direct insult. Without formal education, they might rely on common usage or what they hear from others. If they hear that calling someone a nazi is bad, then using it here would be offensive.  So, the classification might be offensive language because it's using a derogatory term, but if they perceive it as targeting a group with hate, it could be hate speech. But since they don't have the context to know that antifa isn't a fascist group, they might not see the inaccuracy, just the insult. So the rationale would be the words "nazi" and "antifa."
            \textbf{Hate speech}
            \\
      \bottomrule
    \end{tabular}  
  }
  \caption{
    Chain-of-Thought reasoning consistency for HateXplain input text ``\textit{antifa are the real nazis man}'' (GT: \textbf{Normal}) and the \textit{No formal education} persona.
  }
  \label{tab:cot_consistency}
\end{table*}

\subsection{BRWRR (CoS-E and SST-2) Additional Results}
\label{appdix:cose_and_sst}
\begin{figure*}[ht]
  \centering
  \begin{subfigure}[t]{0.48\textwidth}
    \centering
    \includegraphics[width=\linewidth]{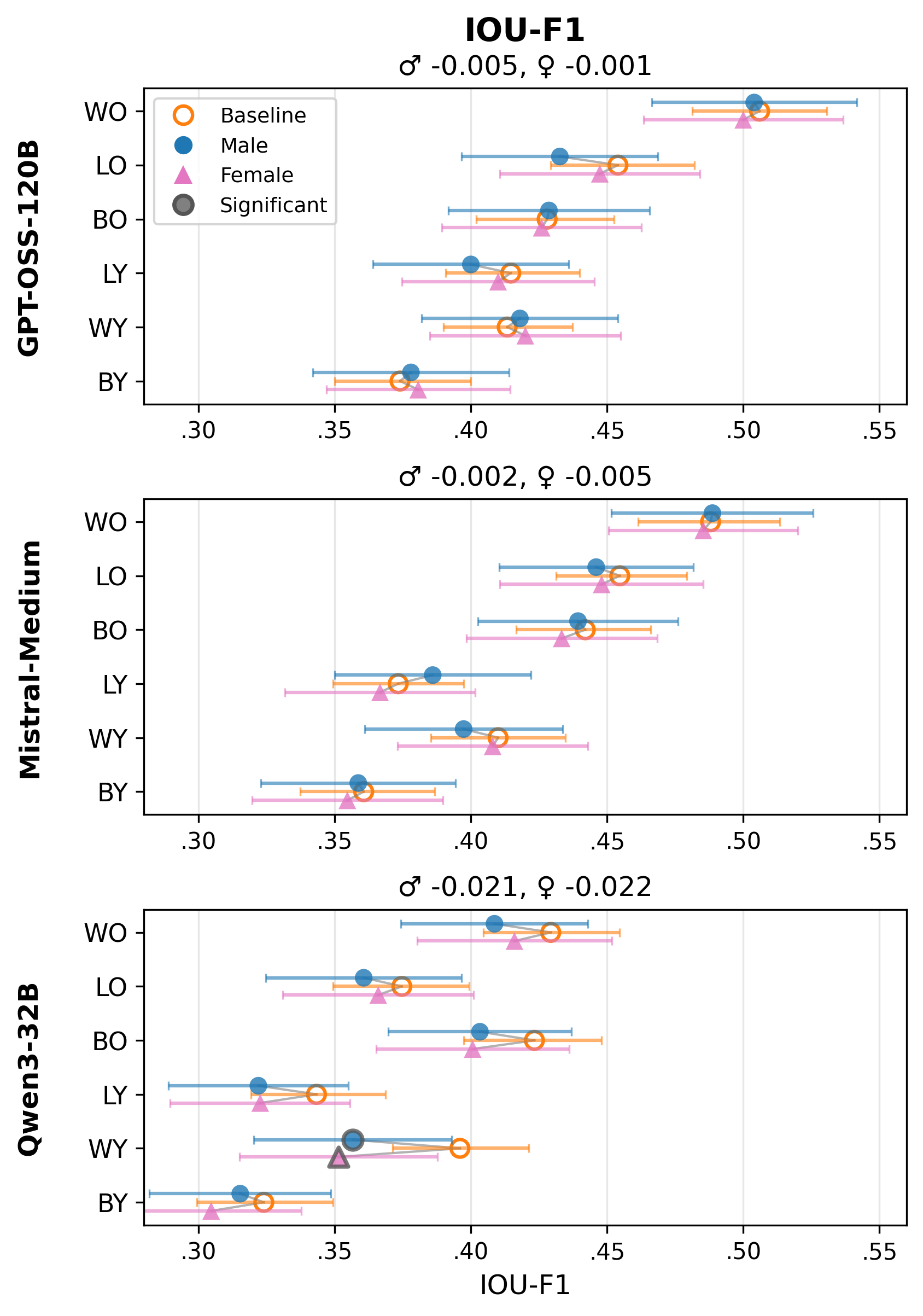}
    \caption{CoS-E subset}
    \label{fig:iou:cose}
  \end{subfigure}\hfill
  \begin{subfigure}[t]{0.48\textwidth}
    \centering
    \includegraphics[width=\linewidth]{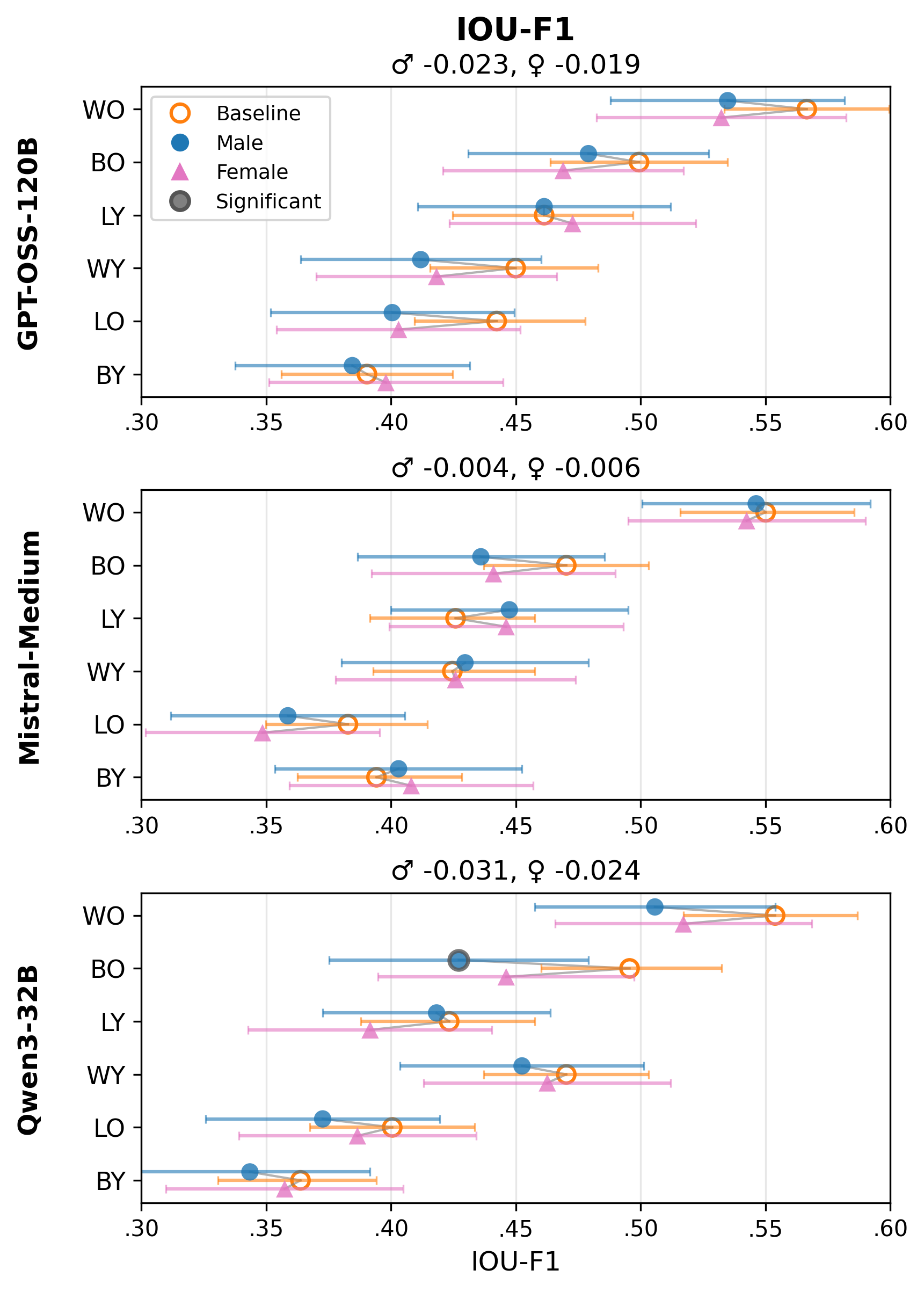}
    \caption{SST-2 subset}
    \label{fig:iou:sst}
  \end{subfigure}
  \caption{Rationale Performance on IOU-$F_1$ for (no persona) baseline and personas across demographic groups. Error bars indicate 95\% CI. BY: African American Young, WY: Caucasian Young, LY: Hispanic Young, BO: African American Old, WO: White Old, LO: Hispanic Old. Each group has its own ground truth labels.}
  \label{fig:iou_baseline_personas_combined}
\end{figure*}

We provide the additional results of IOU-$F_1$ in Figure~\ref{fig:iou_baseline_personas_combined}. We also show the best performing personas for each demographic group's annotations in Table~\ref{tab:best_personas_baseline_cose} (CoS-E) and \ref{tab:best_personas_baseline_sst_f1} (SST-2). The tables show that the best performing personas not always match their corresponding group; nevertheless, the differences between persona (P) and baseline (B) are slight.

\begin{table*}[ht!]
\centering
\adjustbox{width=\textwidth,center}{{%
\begin{tabular}{@{}l*{18}{c}@{}}
\toprule
\multirow{3}{*}{\textbf{Group}} & \multicolumn{6}{c}{\textbf{GPT-OSS-120B}} & \multicolumn{6}{c}{\textbf{Mistral-Medium}} & \multicolumn{6}{c}{\textbf{Qwen3-32B}} \\
\cmidrule(lr){2-7}\cmidrule(lr){8-13}\cmidrule(lr){14-19}
 & \multicolumn{3}{c}{\textbf{Accuracy}} & \multicolumn{3}{c}{\textbf{Token-$F_1$}} & \multicolumn{3}{c}{\textbf{Accuracy}} & \multicolumn{3}{c}{\textbf{Token-$F_1$}} & \multicolumn{3}{c}{\textbf{Accuracy}} & \multicolumn{3}{c}{\textbf{Token-$F_1$}} \\
\cmidrule(lr){2-4}\cmidrule(lr){5-7}\cmidrule(lr){8-10}\cmidrule(lr){11-13}\cmidrule(lr){14-16}\cmidrule(lr){17-19}
 & \textbf{Persona} & \textbf{P} & \textbf{B} & \textbf{Persona} & \textbf{P} & \textbf{B} & \textbf{Persona} & \textbf{P} & \textbf{B} & \textbf{Persona} & \textbf{P} & \textbf{B} & \textbf{Persona} & \textbf{P} & \textbf{B} & \textbf{Persona} & \textbf{P} & \textbf{B} \\
\midrule
\textbf{BO} & LYF & \textbf{76.7} & 75.2 & LYM & \textbf{54.0} & 53.2 & L\textbf{O}M & 75.7 & \textbf{76.5} & WYF & \textbf{52.8} & 51.3 & \textbf{B}YF & 75.5 & \textbf{76.6} & \textbf{B}YM & 48.8 & \textbf{49.2} \\
\textbf{BY} & LOM & \textbf{68.5} & 67.5 & \textbf{B}OM & \textbf{50.7} & 49.9 & W\textbf{Y}M & \textbf{67.7} & 67.5 & LOF & \textbf{49.2} & 48.0 & LOF & \textbf{65.9} & 65.6 & W\textbf{Y}F & 44.5 & \textbf{44.9} \\
\textbf{LO} & BYM & 71.1 & \textbf{71.2} & \textbf{L}\textbf{O}F & \textbf{56.3} & 55.8 & W\textbf{O}F & \textbf{71.1} & 70.7 & W\textbf{O}F & \textbf{55.5} & 54.4 & B\textbf{O}M & \textbf{71.3} & 70.6 & WYM & 49.7 & \textbf{50.0} \\
\textbf{LY} & B\textbf{Y}M & \textbf{67.9} & 67.5 & WOM & \textbf{52.5} & 51.7 & WOF & \textbf{69.9} & 69.2 & W\textbf{Y}F & \textbf{50.2} & 49.2 & \textbf{L}\textbf{Y}M & 67.6 & \textbf{68.5} & \textbf{L}OM & 46.1 & \textbf{46.5} \\
\textbf{WO} & BYM & \textbf{74.8} & 74.4 & LYM & \textbf{59.4} & 58.5 & \textbf{W}\textbf{O}M & 74.3 & \textbf{74.9} & LYM & \textbf{58.4} & 56.9 & L\textbf{O}M & 72.5 & \textbf{73.7} & BYF & 52.7 & \textbf{54.0} \\
\textbf{WY} & B\textbf{Y}M & \textbf{68.1} & 67.1 & \textbf{W}OF & \textbf{54.8} & 53.9 & \textbf{W}OM & \textbf{67.9} & 67.5 & L\textbf{Y}M & \textbf{53.3} & 52.4 & B\textbf{Y}F & 67.0 & \textbf{67.4} & LOM & 49.4 & \textbf{50.9} \\
\bottomrule
\end{tabular}%
}}
\caption{Best Performing Personas vs. Baseline Results on CoS-E (\%). \textbf{P} Persona accuracy, \textbf{B} Baseline accuracy, persona abbreviations: B=African American, W=Caucasian, L=Hispanic, M=Male, F=Female, Y=25 years old, O=45 years old. The differences between persona (P) and baseline (B) are slight.}
\label{tab:best_personas_baseline_cose}
\end{table*}

\begin{table*}[ht!]
\centering
\adjustbox{width=\textwidth,center}{{%
\begin{tabular}{@{}l*{18}{c}@{}}
\toprule
\multirow{3}{*}{\textbf{Group}} & \multicolumn{6}{c}{\textbf{GPT-OSS-120B}} & \multicolumn{6}{c}{\textbf{Mistral-Medium}} & \multicolumn{6}{c}{\textbf{Qwen3-32B}} \\
\cmidrule(lr){2-7}\cmidrule(lr){8-13}\cmidrule(lr){14-19}
 & \multicolumn{3}{c}{\textbf{Binary $F_1$}} & \multicolumn{3}{c}{\textbf{Token-$F_1$}} & \multicolumn{3}{c}{\textbf{Binary $F_1$}} & \multicolumn{3}{c}{\textbf{Token-$F_1$}} & \multicolumn{3}{c}{\textbf{Binary $F_1$}} & \multicolumn{3}{c}{\textbf{Token-$F_1$}} \\
\cmidrule(lr){2-4}\cmidrule(lr){5-7}\cmidrule(lr){8-10}\cmidrule(lr){11-13}\cmidrule(lr){14-16}\cmidrule(lr){17-19}
 & \textbf{Persona} & \textbf{P} & \textbf{B} & \textbf{Persona} & \textbf{P} & \textbf{B} & \textbf{Persona} & \textbf{P} & \textbf{B} & \textbf{Persona} & \textbf{P} & \textbf{B} & \textbf{Persona} & \textbf{P} & \textbf{B} & \textbf{Persona} & \textbf{P} & \textbf{B} \\
\midrule
\textbf{BO} & \textbf{B}\textbf{O}M & \textbf{98.9} & 98.3 & L\textbf{O}F & 60.6 & \textbf{61.1} & W\textbf{O}M & 98.1 & \textbf{98.8} & W\textbf{O}M & 59.7 & \textbf{59.8} & LYF & 97.6 & \textbf{98.0} & WYF & 59.5 & \textbf{61.4} \\
\textbf{BY} & W\textbf{Y}M & \textbf{93.8} & 93.2 & LOF & \textbf{53.3} & 51.8 & WOM & \textbf{93.2} & 93.1 & L\textbf{Y}M & \textbf{52.5} & 51.5 & L\textbf{Y}F & 92.3 & \textbf{93.4} & W\textbf{Y}F & 51.0 & \textbf{51.2} \\
\textbf{LO} & \textbf{L}YF & \textbf{94.8} & 94.1 & W\textbf{O}F & 55.7 & \textbf{56.8} & WYF & 93.9 & \textbf{94.4} & W\textbf{O}F & 54.5 & \textbf{54.8} & WYF & 93.2 & \textbf{93.6} & WYM & \textbf{55.8} & 55.4 \\
\textbf{LY} & \textbf{L}OF & \textbf{97.8} & 97.0 & \textbf{L}OF & \textbf{57.6} & 57.0 & WOF & 97.0 & \textbf{97.2} & \textbf{L}\textbf{Y}M & 56.8 & \textbf{56.9} & W\textbf{Y}F & 95.9 & \textbf{96.4} & W\textbf{Y}M & \textbf{57.2} & 57.1 \\
\textbf{WO} & \textbf{W}\textbf{O}F & \textbf{97.7} & 97.0 & \textbf{W}YM & 64.7 & \textbf{65.0} & L\textbf{O}F & 96.9 & \textbf{97.3} & LYM & 63.6 & \textbf{63.9} & \textbf{W}YF & 95.8 & \textbf{96.5} & \textbf{W}YM & 63.4 & \textbf{63.9} \\
\textbf{WY} & BOF & \textbf{95.1} & 94.4 & \textbf{W}OF & 57.6 & \textbf{58.4} & LOF & 94.3 & \textbf{95.0} & \textbf{W}OM & 57.6 & \textbf{58.1} & L\textbf{Y}M & 94.2 & \textbf{94.4} & \textbf{W}\textbf{Y}M & 59.3 & \textbf{59.6} \\
\bottomrule
\end{tabular}%
}}
\caption{Best Performing Personas vs. Baseline Results on SST-2 (\%). Persona abbreviations: B=African American, W=Caucasian, L=Hispanic, M=Male, F=Female, Y=25 years old, O=45 years old. The differences between persona (P) and baseline (B) are slight.}
\label{tab:best_personas_baseline_sst_f1}
\end{table*}

\end{document}